\documentclass{article}

\usepackage[eandd, final]{neurips_2026}

\usepackage[utf8]{inputenc} 
\usepackage[T1]{fontenc}    
\usepackage{hyperref}       
\usepackage{url}            
\usepackage{booktabs}       
\usepackage{amsfonts}       
\usepackage{nicefrac}       
\usepackage{microtype}      
\usepackage{xcolor}         
\usepackage{multirow}
\usepackage{graphicx}
\usepackage{subcaption}
\usepackage[table]{xcolor}  
\usepackage{soul}
\usepackage{amsmath, amssymb, amsfonts, bm}
\usepackage{booktabs}
\usepackage{algorithm}
\usepackage{tikz}
\usetikzlibrary{arrows.meta, positioning, shapes.geometric, fit}
\usepackage{algpseudocode}
\usepackage{titletoc}
\usepackage{iftex}
\ifluatex
  \usepackage{emoji}
\else
  \newcommand{\emoji}[1]{}
\fi

\titlecontents{section}
  [2em]                  
  {\addvspace{8pt}}
  {\contentslabel{2em}}  
  {}
  {\titlerule*[0.6pc]{.}\contentspage}

\titlecontents{subsection}
  [4.2em]                
  {}
  {\contentslabel{2.8em}}
  {}
  {\titlerule*[0.6pc]{.}\contentspage}

\title{CORAL: A Benchmark for Structure-aware and Brain-wide Neuron Reconstruction in Light Microscopy}

\author{
  Zekang Yang$^{1,2}$, Jiamin Li$^{2,3,4}$, Zhenghua Li$^{1,2}$, Jiaqi Fan$^{2,3,4}$,\\ \textbf{Zengcai Guo$^{2,3,4*}$, Xiaolin Hu$^{1,2,5}$\thanks{Corresponding authors.}}\\
  $^{1}$Department of Computer Science and Technology, \\
  Institute for AI, BNRist, Tsinghua University, Beijing 100084, China\\
  $^{2}$IDG/McGovern Institute for Brain Research, Tsinghua University,
Beijing 100084, China\\
  $^{3}$School of Basic Medical Sciences, Tsinghua University, Beijing, 100084, China\\
  $^{4}$Tsinghua-Peking Center for Life Sciences, Beijing, 100084, China\\
  $^{5}$Chinese Institute for Brain Research (CIBR), Beijing 100010, China\\
  \{yzk25, li-zh24, jq-fan24\}@mails.tsinghua.edu.cn, li-jm20@tsinghua.org.cn,\\
  \{guozengcai,xlhu\}@tsinghua.edu.cn
}

\begin{document}
\maketitle

\begin{abstract}
Automatic neuron reconstruction from light microscopy images is a central problem in computational neuroanatomy.
While recent methods have achieved encouraging results on local image blocks, it remains unclear whether such progress translates to reconstruction that is both structurally accurate and scalable to the whole-brain scale.
We present CORAL, the first benchmark for structure-aware evaluation of automatic neuron reconstruction from light microscopy images at both local and whole-brain scales.
Built on a high-quality whole-brain fMOST dataset with carefully curated annotations, CORAL establishes two progressive tasks: block-level reconstruction, which evaluates reconstruction methods under limited spatial context, and brain-wide reconstruction, which assesses complete neuron reconstruction at the whole-brain scale.
To account for topological correctness beyond geometric distance similarity, we introduce a structure-aware metric based on fiber prediction.
To further achieve complete neuron reconstruction across the entire brain, we develop a brain-wide neuron tracing framework that extends arbitrary local reconstruction methods to the whole-brain scale through an iterative local-to-global process.
Using this benchmark, we provide the first structure-aware comparison of mainstream methods for local neuron reconstruction and further evaluate their performance in brain-wide reconstruction.
Our results underscore the importance of structure-aware evaluation and the need for more robust methods for complete neuron reconstruction.
All datasets and code are available at \href{https://github.com/yang-ze-kang/CORAL}{https://github.com/yang-ze-kang/CORAL}.
\end{abstract}
\section{Introduction}

Reconstructing neuronal morphologies from light microscopy images is a fundamental problem in computational neuroanatomy, with broad importance for analyzing neuronal cell types, long-range projections, and large-scale brain connectivity~\cite{winnubst2019reconstruction, liu2025connectivity, gao2026integrative}. Over the past decade, advances in image enhancement, 3D segmentation, and tracing algorithms have substantially improved the accuracy and scalability of automatic neuron reconstruction~\cite{APP2, neuTube, ADTL-Net}. As a result, modern methods can already achieve promising geometric accuracy on local image blocks.

Despite these advances, a fundamental question remains largely unanswered: \emph{how far are current automatic methods from achieving structurally accurate neuron reconstruction at the whole-brain scale?} Existing public benchmarks~\cite{DIADEM, BigNeuron} mainly evaluate reconstruction performance on local image blocks and isolated neuronal structures. 
However, in large-scale fMOST imaging data, many neurons span distances far beyond a single local volume, meaning that only partial neuronal structures are visible within any given local image.
As a result, these benchmarks provide limited insight into whole-brain reconstruction performance and thus fail to capture its central challenges, including global topological correctness and long-range continuity preservation.

A second limitation of existing evaluation protocols is that many commonly used metrics are primarily geometric rather than structure-aware. Segmentation metrics, such as Dice-style overlap measures, quantify foreground agreement at the voxel level, while geometry-based metrics evaluate spatial position or connectivity consistency between prediction and reference \cite{Vaa3D,PyNeval}.
However, neuron reconstruction is fundamentally a tree structure prediction problem: a reconstruction can still be biologically incorrect even when most points are well aligned, if a small number of connectivity errors alter the branching structure or long-range trajectories of the neuronal tree.
In particular, root-to-leaf fibers are often more critical than local geometric similarity for assessing whether a reconstruction preserves the correct neuronal organization. This calls for a benchmark that evaluates not only local geometric fidelity, but also the correctness of global neuronal structure.

To address these limitations, we present CORAL, a benchmark for structure-aware evaluation
of automatic neuron reconstruction from light microscopy images at both local
and whole-brain scales. CORAL is built on a high-quality whole-brain fMOST dataset with carefully curated neuron annotations and supports evaluation under two progressive settings: \emph{block-level reconstruction}, which is designed to develop reconstruction algorithms, and \emph{brain-wide reconstruction}, which evaluates complete neuron reconstruction at the whole-brain scale. To better assess reconstruction quality, we further introduce structure-aware metrics based on \emph{keypoint prediction} and \emph{fiber prediction}, which explicitly measure the recovery of bifurcation/termination events and root-to-leaf neuronal fibers. In addition, to enable standardized large-scale evaluation of existing reconstruction methods in the brain-wide setting, we develop a whole-brain tracing framework that extends arbitrary local reconstruction algorithms to the whole-brain scale.

Using this benchmark, we conducted a standardized large-scale evaluation of representative reconstruction methods under both block-level and brain-wide settings.
Our experiments yield three main findings.
First, in two-stage reconstruction methods, the segmentation method and the reconstruction method are coupled, making algorithm optimization more challenging.
Second, geometry-based metrics are insufficient to characterize final reconstruction quality: minor local errors can lead to global topological failures, whereas structure-aware keypoint- and fiber-based metrics provide a substantially more comprehensive assessment of reconstruction quality.
Third, current methods remain far from achieving structurally correct reconstruction and fully automated brain-wide reconstruction. Even small local structural errors can propagate into large-scale errors across the whole brain, imposing higher requirements on automated reconstruction methods.
These findings demonstrated that CORAL provides a rigorous testbed for developing reconstruction methods and evaluating brain-wide neuron reconstruction.
\section{Related Work}

\subsection{Datasets}

Existing public datasets for automated neuron reconstruction are predominantly constructed from small-scale volumetric image blocks. The DIADEM~\cite{DIADEM} dataset is one of the earliest standardized benchmarks, providing manually annotated neuron reconstructions across several imaging conditions.
The BigNeuron~\cite{BigNeuron} project expands the diversity with collections, yet still limited to small-volume image blocks.
More recent datasets, including  CWMBS~\cite{ADTL-Net} and NeuroFly~\cite{NeuroFly}, further increase variability in morphology and imaging modalities, but are likewise derived from cropped sub-volumes.
The Brain Image Library (BIL)~\cite{bil} hosts a large collection of whole-brain image datasets and neuron annotations, supporting neuroscientists in tracing neuronal circuits and comprehensively map cell types.
However, due to the lack of a standardized evaluation protocol for automatic reconstruction algorithms, these data have not been widely used in the automatic neuron reconstruction community. 
To our knowledge, only Liu et al.~\cite{ADTL-Net} have evaluated reconstruction performance on four complete neurons selected from BIL. Nevertheless, their study did not establish a standard evaluation protocol.

\subsection{Methods}
Neuron reconstruction has been studied for several decades.
Early algorithms primarily relied on conventional computational techniques, such as path pruning~\cite{APP, APP2}, graph search~\cite{Kimimaro, SmartTracing}, active contours~\cite{wang2011broadly}, and hand-crafted tubular-structure cues~\cite{neuTube}.
With the advent of deep learning, researchers have increasingly used deep models to segment and enhance neuronal structures before applying traditional reconstruction algorithms, achieving better performance than directly applying reconstruction methods~\cite{li2017deep, wang2019segmenting}. Subsequent improvements have mainly focused on strengthening segmentation models, including improving foreground prediction accuracy~\cite{yang2020neuron, ADTL-Net} and preserving the continuity of linear structures~\cite{yan2025glancing}.
In recent years, some researchers have sought to develop deep learning-based reconstruction methods by predicting neuronal orientations~\cite{SPE-DNR} or the locations of subsequent tracing points~\cite{NetTracer}.

\subsection{Metrics}
\begin{figure}[htbp]
    \includegraphics[width=0.98\linewidth]{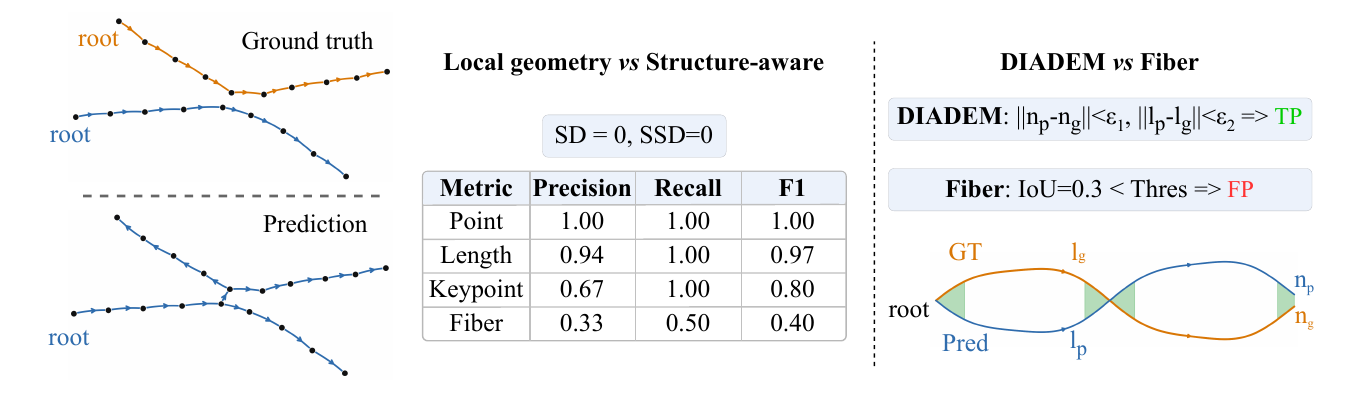}
    \caption{Comparison of different metrics. The left uses a representative prediction error to highlight the limitations of geometry-based metrics. The right illustrates the limitations of DIADEM.}\label{fig:comp_metric}
\end{figure}

A diverse set of metrics has been proposed to evaluate neuronal reconstruction from multiple perspectives.
From a segmentation viewpoint, voxel-wise metrics such as F1-score and Dice coefficient are commonly used to measure the overlap between predicted neuronal regions and ground truth, reflecting foreground classification accuracy.
Metrics designed for neuron reconstruction can be categorized into two following groups:

\paragraph{Geometry-based metrics.}
Peng et al.~\cite{peng2010automatic} introduced distance-based metrics that measure the minimum spatial distance between reconstructed and ground-truth structures, including spatial distance (SD) and significant spatial distance (SSD), which are the most commonly used metrics in neuron reconstruction evaluation.
Liu et al.~\cite{ADTL-Net} treated predicted and ground-truth point pairs whose distance is below a predefined threshold as true positives, and accordingly define precision, recall, F1 score (referred to as Point-F1 for simplicity), and Miss-Extra Scores (MES).
Zhang et al.~\cite{PyNeval} further consider point connectivity by treating an edge pair as a true positive when both its endpoints and length are matched, and propose the Length-F1 metric to evaluate predicted-edge accuracy.
\textbf{Limitations:} These metrics assess reconstruction quality only from a local geometric perspective, whereas the correctness of the topological structure is essential and cannot be overlooked.
The left panel of Figure~\ref{fig:comp_metric} illustrates an example showing why structure-aware is essential.
The prediction contains the same number of nodes and spatial locations with the ground truth, causing SD, SSD, and Point-F1 to collapse.
The prediction traces is routed onto a neighboring fiber and occurs a merging error, leaving only one fiber correctly recovered. Locally, this corresponds to only one incorrectly predicted edge, resulting in Length-F1 having an incorrect high value.

\paragraph{Structure-aware metrics.}
Feng et al.~\cite{neuTube} adopted the critical node metric to evaluate the prediction accuracy at key locations (bifurcations and leafs) that affect structural correctness.
Tree edit distance (TED)~\cite{TED} has also been introduced to evaluate neuron reconstruction quality.
DIADEM metric~\cite{DIADEMMetric} further extends TED by checking whether each matched node has a corresponding matched ancestor and whether the difference between their respective path lengths to the ancestor is below a predefined threshold.
\textbf{Limitations:} DIADEM has several design limitations that have prevented its widespread adoption: (1) It relies on a complex computation protocol and produces only an abstract score, making it less suitable for inspecting bad cases or guiding algorithm improvement. (2) It relies on a directed tree structure rooted at a known soma location, which makes it unsuitable as a block-level metric, where local image blocks often lack the soma and predicted reconstructions are collections of undirected graphs. (3) As shown on the right side of Figure~\ref{fig:comp_metric}, DIADEM still has a design limitation. In practical tracing, dense neuronal fibers or merging errors may allow a matched node to reach the matched ancestor node through a different path with the same path length, causing DIADEM to spuriously count it as a correct prediction.

\section{Dataset Description}

\subsection{Data Curation and Annotation Protocol}
Our mouse brain samples underwent a standard tissue preparation protocol for fMOST imaging, including dehydration, resin infiltration, embedding and imaging~\cite{gang2017embedding, fMOST}.
The acquired raw image tiles were subjected to a series of preprocessing steps, comprising pixel-wise dual-channel registration, image stitching, artifact removal, and signal enhancement.
The in-plane resolution is $0.35 \mu$m in the $x$–$y$ axes, and the axial resolution is $1 \mu$m along the $z$ axis.
We will also upload our mouse brain images to the Brain Image Library~\cite{bil}.

To facilitate the development and evaluation of automatic reconstruction algorithms, we selected a dataset containing 32 neurons for annotation.
For quality control, we established a three-stage annotation protocol involving three independent annotators.
First, neuronal fibers were annotated by the first annotator.
Next, a different annotator independently reviewed all reconstructed fibers from scratch; any missing branches or incomplete structures required re-annotation.
Finally, a third annotator performed global cross-validation.
All annotation and verification were performed using Fast Neuron Tracing (FNT) software~\cite{FNT} and were exported in the standardized SWC format~\cite{NeuroMorpho}.
Compared to existing datasets, our dataset is, to the best of our knowledge, the first to achieve near-exhaustive annotation of all visible neurons within an entire brain volume.

\begin{figure}[htbp]
    \centering
    \begin{minipage}[c]{0.45\linewidth}
        \centering
        \includegraphics[width=\linewidth]{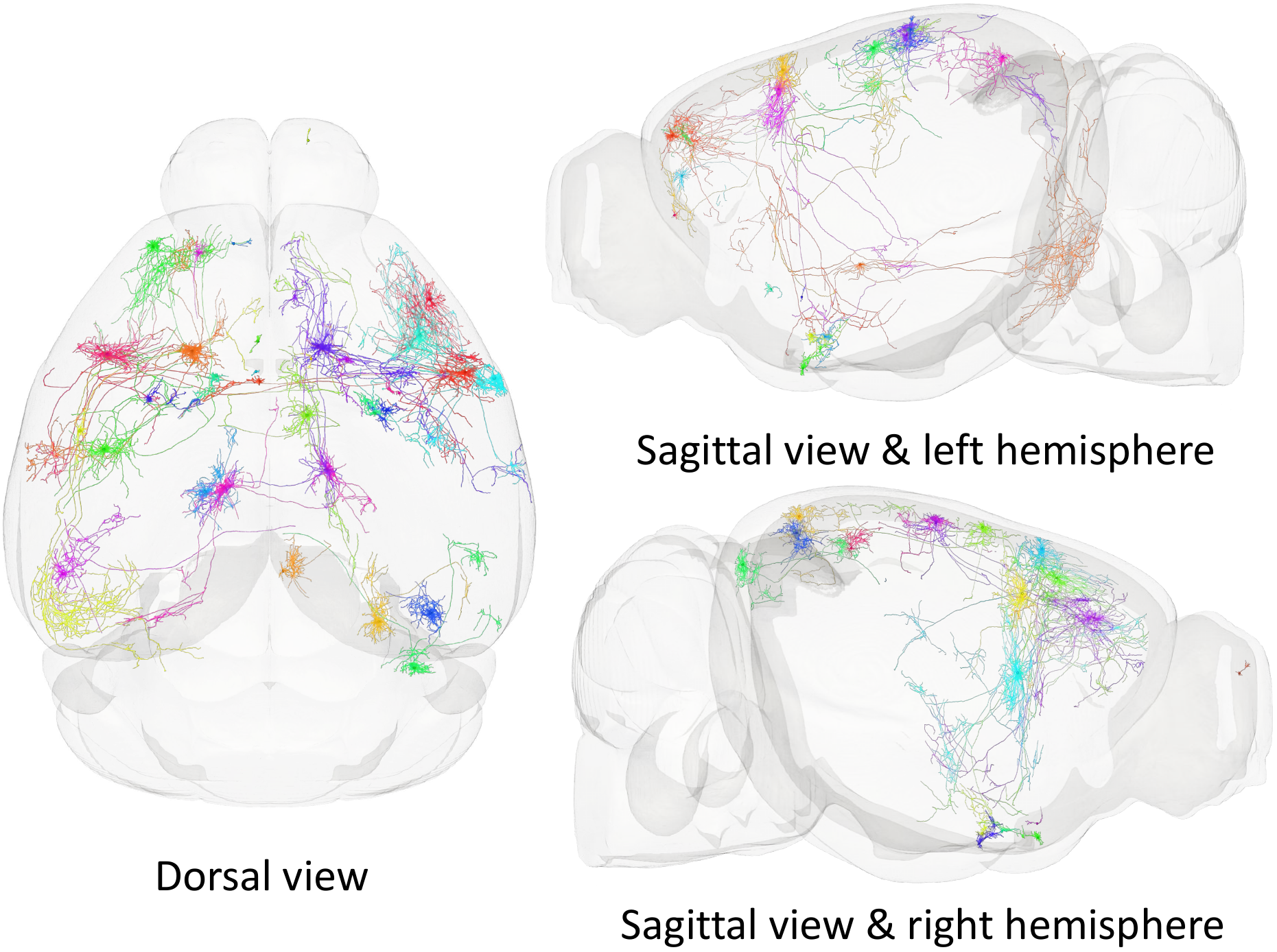}
        \subcaption{Spatial distribution of 32 neurons.}
        \label{fig:visual_neuron_spatial}
    \end{minipage}
    \hfill
    \begin{minipage}[c]{0.54\linewidth}
        \centering
        \begin{minipage}[c]{\linewidth}
            \centering
            \includegraphics[width=\linewidth]{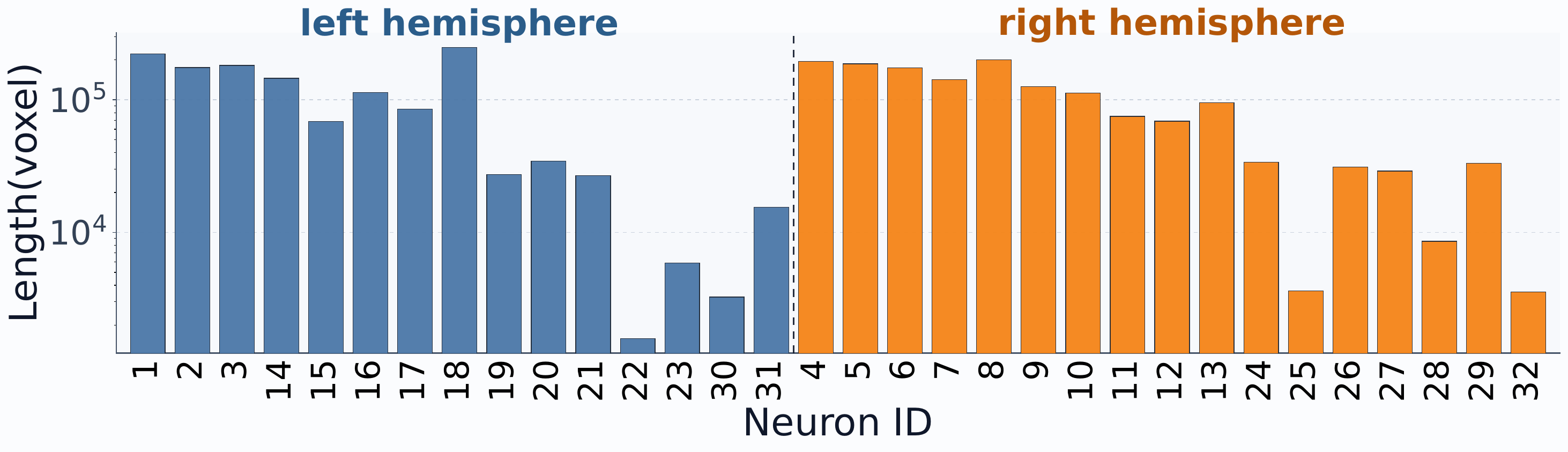}
            \subcaption{Lengths of neurons}
            \label{fig:visual_neuron_length}
        \end{minipage}
        \vspace{0.5em}
        \begin{minipage}[c]{\linewidth}
            \centering
            \includegraphics[width=\linewidth]{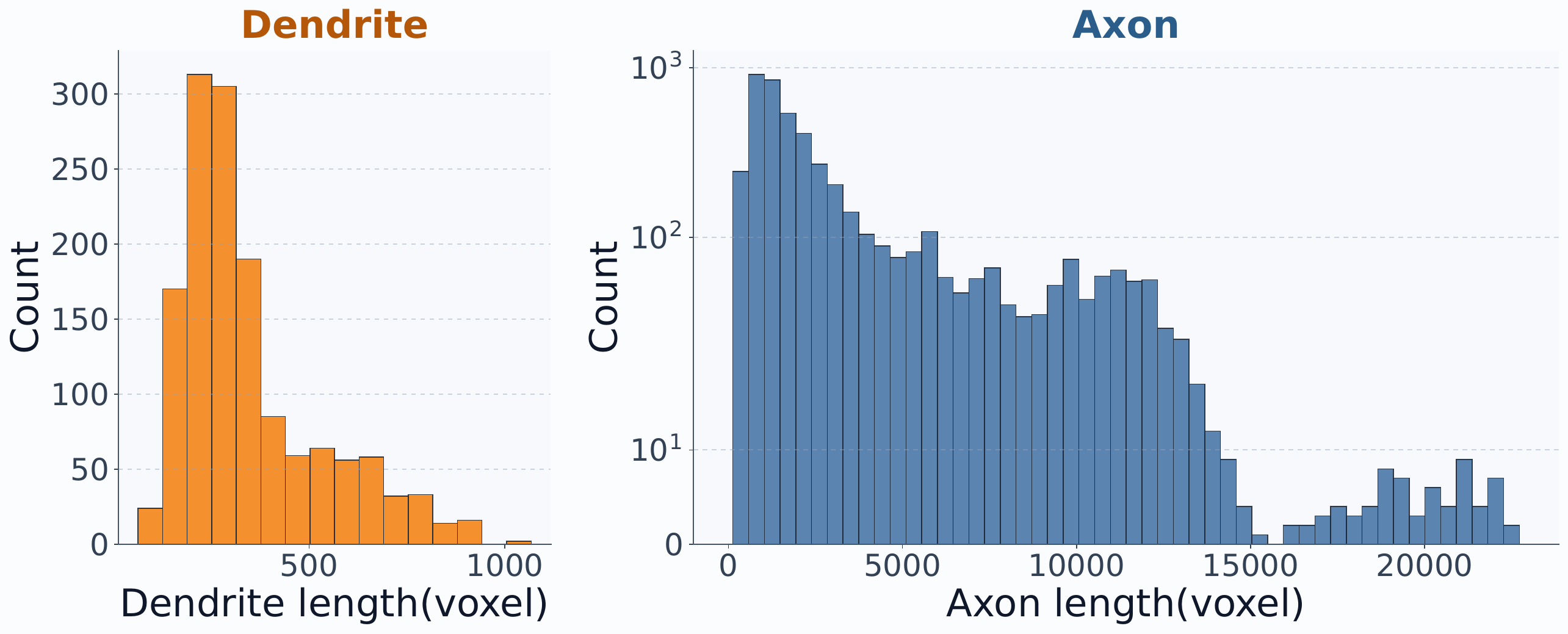}
            \subcaption{Distribution of fiber lengths.}
            \label{fig:visual_neuron_fiber}
        \end{minipage}
    \end{minipage}
    \caption{Visualization of dataset statistics. (a) Spatial distribution of the 32 neurons in the mouse brain from multiple views. (b) Total length of each neuron, with blue denoting neurons whose somata are located in the left hemisphere and orange denoting those whose somata are located in the right hemisphere. (c) Length distribution of dendrites (orange) and axons (blue).}
    \label{fig:visual_neuron}
\end{figure}

\subsection{Dataset Statistics}
Our dataset includes 32 neurons with complete morphological reconstructions, among which 15 have their somata located in the left hemisphere and 17 in the right hemisphere.
The neurons are sparsely distributed across the cerebral cortex, as shown in Figure~\ref{fig:visual_neuron_spatial}.
Figure~\ref{fig:visual_neuron_length} shows the total length of each reconstructed neuron. The shortest neuron spans 695~$\mu$m (1,584 voxels), whereas the longest reaches 110,292~$\mu$m (148,842 voxels), with neurons of varying sizes distributed across both the left and right hemispheres.
Figure~\ref{fig:visual_neuron_fiber} further shows the length distributions of dendrites and axons. In total, the dataset contains 1,421 dendritic fibers and 4,931 axonal fibers. The dendrite lengths range from 33 to 429$\mu$m (62–1,067 voxels), whereas axon lengths exhibit a much broader range, from 103 to 9,643 $\mu$m (273–22,720 voxels), highlighting the challenge of preserving long-range continuity during axon tracing.
\section{CORAL Benchmark}
\begin{figure}[htbp]
    \centering
    \includegraphics[width=0.98\linewidth]{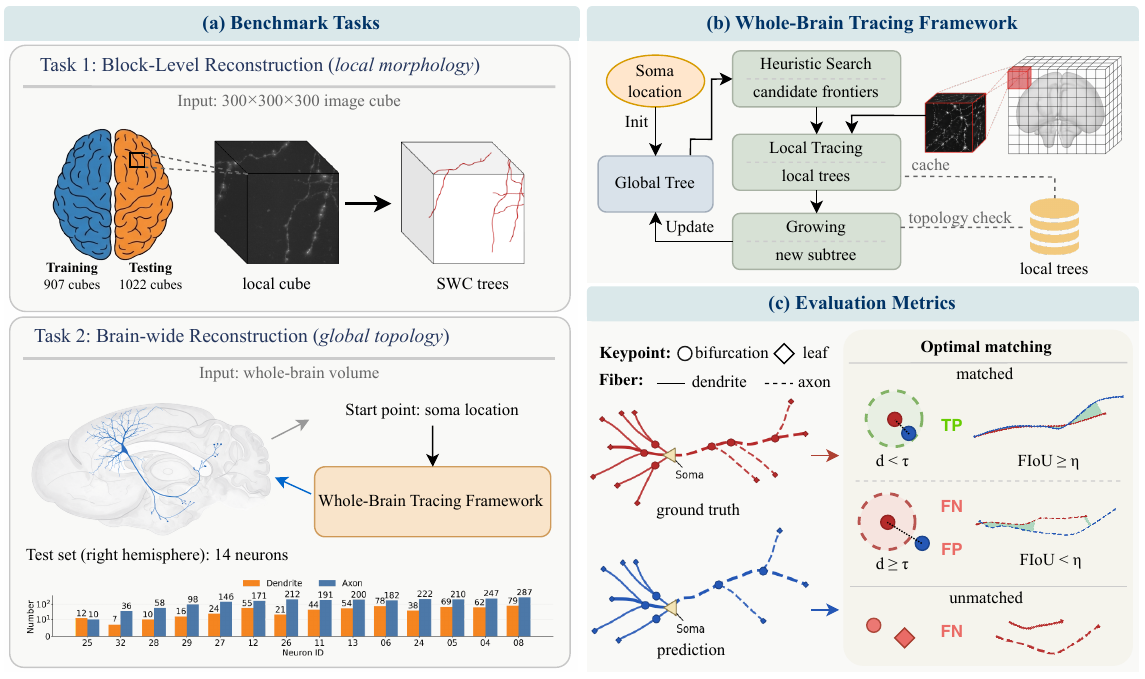}
    \caption{Overview of the CORAL benchmark. (a) shows the two benchmark tasks. (b) shows the whole-brain tracing framework. (c) shows the two structure-aware metrics (Keypoint and Fiber), where $d$ denotes the distance between matched keypoints and $\tau$ is the corresponding distance threshold; The green region indicates the overlapping area; FIoU denotes the fiber intersection-over-union of overlapped lengths between matched fibers, and $\eta$ is the FIoU threshold for valid matching.}
    \label{fig:overview}
\end{figure}

Figure~\ref{fig:overview} provides an overview of CORAL, whose ultimate goal is to evaluate the performance of algorithms in brain-wide neuron reconstruction.
Due to the high resolution of brain images, directly importing the whole volume for brain-wide neuron reconstruction is impractical.
To address this challenge, we developed a whole-brain tracing framework (WBTF) (Sec~\ref{sec:method_wbtf}) that can equip arbitrary reconstruction algorithms with the ability to reconstruct brain-wide neurons.
CORAL defines two progressive tasks: block-level reconstruction and brain-wide reconstruction (Sec~\ref{sec:method_task}). 
The diverse block-level images, together with the WBTF, enable reconstruction algorithms to be developed and trained on block images and then deployed for brain-wide tracing inference.
Moreover, we design the Fiber metric, a more comprehensive yet simple and intuitive metric, to evaluate the performance of different algorithms (Sec~\ref{sec:method_metric}).


\subsection{Whole-brain tracing framework}
\label{sec:method_wbtf}
As shown in Figure~\ref{fig:overview}, the proposed whole-brain tracing framework consists of three main components: heuristic search, local tracing, and growing.
First, for heuristic search, we partition the high-resolution whole-brain volume into a 3D grid. 
Each grid position corresponds to a block image centered at that position, and block images associated with neighboring grid positions overlap with each other.
When tracing a neuron, the algorithm starts from the grid position containing the soma.
It then selects leaf nodes, as well as nodes close to the boundary of the already traced region, from the current reconstruction as candidate search locations.
By progressively moving the field of view, the framework expands the reconstruction and obtains an intermediate result over the brain-wide.
Second, for local tracing, given the block image corresponding to the current grid point, any neuron reconstruction algorithm can be applied to obtain a local reconstruction result.
Third, for growing, the framework compares the local reconstruction in the current field of view with the overlapping part of the existing global reconstruction, and stitches newly traced subtrees from the current view into the global reconstruction.
See implementation details in the appendix~\ref{sec:whole_brain_tracing_framework}.

\subsection{Task Settings}
\label{sec:method_task}
We use a hemisphere-based partition throughout the benchmark: data from the left hemisphere are used for training, while data from the right hemisphere are reserved for testing.
This setting prevents data leakage during testing.

\paragraph{Block-level reconstruction.}
Considering the diversity of neuronal morphology, sparsity, and signal intensity across the whole brain, we cropped 1,937 diverse blocks from the entire brain volume for training and testing different reconstruction methods.
Each block has the size of $300 \times 300 \times 300$ and paired with its corresponding SWC annotation.
We assigned 907 left-hemisphere blocks to the training set and 1,022 right-hemisphere blocks to the test set, while holding out 8 blocks spanning both hemispheres.
Compared with previous block-level datasets~\cite{ADTL-Net}, our block-level dataset is ten times larger and captures greater diversity in neuronal structures and local imaging environments, making it better suited for training reconstruction algorithms.

\paragraph{Brain-wide reconstruction.}
In this task, algorithms start from the initial tree at the soma location of each neuron and reconstruct the complete neuron across the brain-wide.
Among the 17 neurons in the right hemisphere, three extend into the left hemisphere. To prevent data leakage, we selected the remaining 14 neurons as the test set.
Among these 14 neurons, the shortest neuron to be reconstructed has a total length of 1,404~$\mu$m (3,560 voxels), whereas the longest reaches 86,680~$\mu$m (200,067 voxels).
In total, 746 dendrites and 2785 axons are required to be reconstructed.
Figure~\ref{fig:overview} shows the number of axons and dendrites to be reconstructed for each neuron.

\subsection{Evaluation Metrics}
\label{sec:method_metric}
We adopt two structure-aware metrics, Keypoint metric and Fiber metric, to evaluate reconstruction quality.
The Keypoint metric measures how accurately an algorithm predicts critical locations.
Unlike previous metrics~\cite{neuTube}, we separately evaluate the reconstruction accuracy of bifurcations and leafs, as they affect reconstruction quality through different patterns.

Inspired by the primary goal of neuron reconstruction, which is to trace all neurites (dendrites and axons) starting from the soma, as Figure~\ref{fig:overview}c shows, we design the Fiber metric to directly measure the reconstruction performance of each neurite.
Given the ground truth tree $T_g$ and the predicted tree $T_p$, we first decompose them into sets of fibers $\mathcal{F}_g=\{f_1, f_2, \dots, f_m\}$ and $\mathcal{F}_p=\{f'_1, f'_2, \dots, f'_n\}$, where each fiber corresponds to a root-to-leaf path in the tree. 
Each fiber is a set of continuous edges, let $\mathcal{C}_f$ denotes the midpoints of edges in fiber $f$, and $\delta_c$ denotes the length of the edge corresponding to midpoint $c$.
To quantify the geometric similarity between two fibers, we define the fiber overlap as the sum of the lengths of edges whose corresponding midpoints in the ground-truth and predicted fibers are within a distance threshold $\tau$:
\[
    \mathrm{Overlap}(f_g, f_p)
    = \frac{1}{2}(\sum_{\mathbf{c} \in \mathcal{C}_g}
    \mathbb{I}\big( \mathrm{dist}(\mathbf{c}, f_p) \leq \tau \big) \cdot \delta_c + \sum_{\mathbf{c} \in \mathcal{C}_p}
    \mathbb{I}\big( \mathrm{dist}(\mathbf{c}, f_g) \leq \tau \big) \cdot \delta_c),
\]
The green region in Figure~\ref{fig:overview}c represents the overlap region.
We then can define \emph{fiber intersection-over-union (FIoU)} based on fiber overlap:
\[
    \mathrm{FIoU}(f_g, f_p) = \frac{\mathrm{Overlap}(f_g, f_p)}{\mathrm{Length}(f_g) + \mathrm{Length}(f_p) - \mathrm{Overlap}(f_g, f_p)},
\]
where $Length(f)$ denotes the total length of fiber $f$.
Then we compute the FIoU between each pair of predicted and ground truth fibers, and perform optimal matching based on the FIoU to identify matched fiber pairs $\mathcal{M}$.
The unmatched fibers in $\mathcal{F}_g$ and $\mathcal{F}_p$ are considered false negatives (FN) and false positives (FP), respectively.
The fiber pair is considered a true positives (TP) only if its FIoU exceeds the threshold $\eta$ :
\[
    \mathrm{TP} = \sum_{(f_g,f_p)\in\mathcal{M}}
    \mathbb{I}\big(\mathrm{FIoU}(f_g, f_p)\geq \eta\big),
\]
Conversely, matched fiber pairs with an FIoU below the threshold $\eta$ are also counted as false positives and false negatives.
Then we have the definitions of precision, recall, and F1 for Fiber metric:
\[
    \mathrm{Precision}=\frac{\mathrm{TP}}{|\mathcal{F}^{\mathrm{pred}}|},\qquad
    \mathrm{Recall}=\frac{\mathrm{TP}}{|\mathcal{F}^{\mathrm{gt}}|},\qquad
    \mathrm{F1}=\frac{2*\mathrm{Precision}*\mathrm{Recall}}{\mathrm{Precision}+\mathrm{Recall}}.
\]
As shown in Figure~\ref{fig:comp_metric}, the Fiber metric provides a more accurate characterization of reconstruction quality. Compared with DIADEM, it is more intuitive and flexible, making it a comprehensive structure-aware metric.
\section{Experiments and Results}

\subsection{Block-level Reconstruction}
Recent high-performing neuron reconstruction algorithms mainly adopt a two-stage pipeline consisting of a segmentation model followed by a reconstruction method.
Therefore, we first selected ten segmentation models,  including state-of-the-art 3D segmentation models~\cite{VNet, nnU-Net, UNETR, SwinUNETR, MedNeXt, xing2025segmamba} as well as models specifically designed for curvilinear-structure neuron segmentation~\cite{IVNet, clDice, DSCNet, ADTL-Net}.
We then combined these segmentation models with different reconstruction methods~\cite{APP2,SmartTracing, neuTube, Kimimaro}.
In addition, we compared two learning-based reconstruction methods~\cite{SPE-DNR, NetTracer}.

\begin{figure}[htbp]
    \centering
    \begin{minipage}[c]{0.42\textwidth}
        \centering
        \captionof{table}{Average speed of different reconstruction methods.}
        \label{tab:avg_speed}
        \begin{tabular}{cc}
            \toprule
            Method & Speed (s/cube) \\
            \midrule
            APP2~\cite{APP2}       & 192.857 \\
            SmartTrace~\cite{SmartTracing} & 155.760 \\
            neuTube~\cite{neuTube}    & 12.032 \\
            Kimimaro~\cite{Kimimaro}   & 0.891 \\
            SPE-DNR~\cite{SPE-DNR}    & 12.005 \\
            NETracer~\cite{NetTracer}  & 1025.043 \\
            \bottomrule
        \end{tabular}
    \end{minipage}
    \hspace{0.05\textwidth}
    \begin{minipage}[c]{0.42\textwidth}
        \centering
        \includegraphics[width=0.8\textwidth]{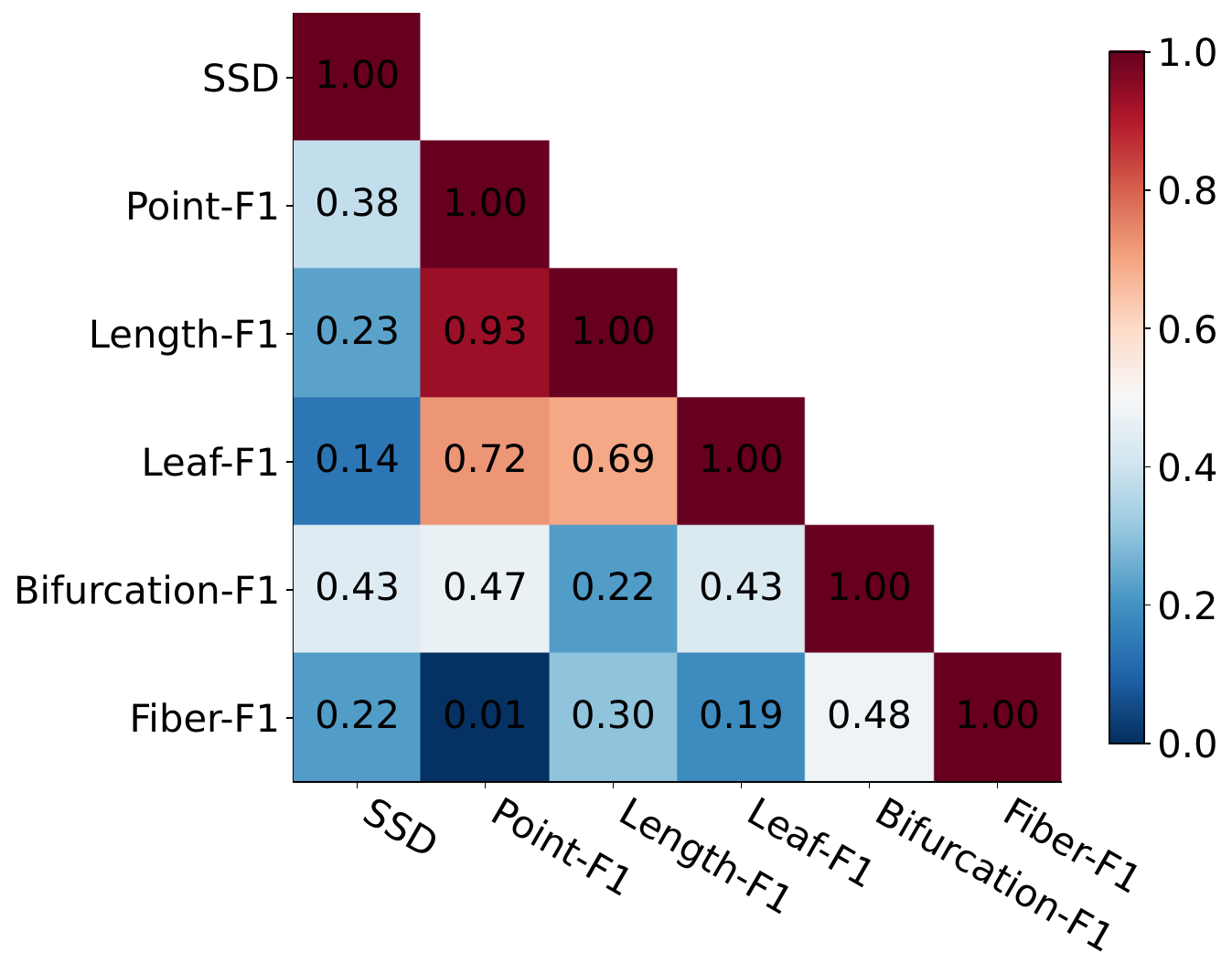}
        \caption{Pearson correlation of metrics.}
        \label{fig:res_compare_metrics}
    \end{minipage}
    \vspace{-4mm}
\end{figure}

The segmentation results are reported in Appendix Table~\ref{tab:block_seg}, showing that SegMamba and IVNet achieve the best segmentation performance.
Figure~\ref{fig:res_trace} shows the reconstruction results produced by different methods and presents commonly used metrics with our recommended metrics, showing that the combination of SegMamba and neuTube achieves the best structure-aware reconstruction quality (Fiber-F1: 46.8).
Table~\ref{tab:avg_speed} shows the speeds of different reconstruction methods.
Different methods exhibit large differences in reconstruction speed and can be roughly divided into three orders of magnitude. Kimimaro is the fastest, followed by neuTube and SPE-DNR; APP2 and SmartTrace are slower, while NETracer is the slowest.
Detailed experimental results are provided in the Appendix.
From the experimental results, we draw the following three main findings:

\begin{figure}[htbp]
    \centering
    \includegraphics[width=\linewidth]{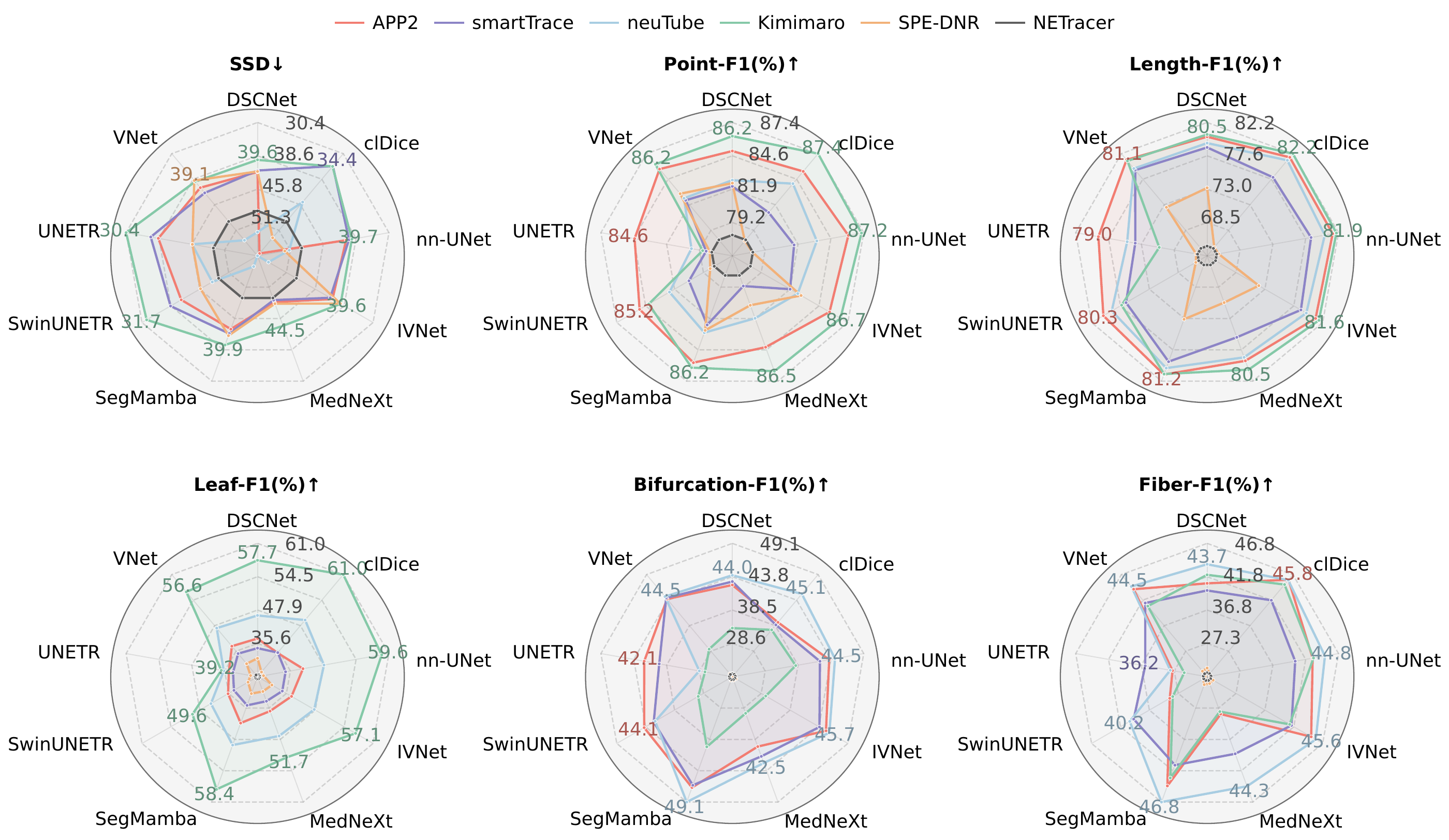}
    \caption{Results of different reconstruction methods on the block-level task. The top three metrics (SSD, Point-F1, Length-F1) are geometry-based, whereas the bottom three metrics (Leaf-F1, Bifurcation-F1, Fiber-F1) are structure-aware metrics that we recommend for evaluating neuron reconstruction quality. In each radar plot, the axes denote the segmentation methods, and the closed curves correspond to the reconstruction methods. NETracer does not rely on segmentation methods and therefore forms an approximately circular curve.}
    \label{fig:res_trace}
\end{figure}
\paragraph{In two-stage reconstruction methods, the segmentation method and the reconstruction method are coupled.}
Previous studies have shown that traditional reconstruction methods can achieve better performance when preceded by segmentation preprocessing.
However, better segmentation performance does not imply better reconstruction quality: clDice performed substantially worse than SegMamba and IVNet in segmentation performance (Dice: 48.04 vs. 54.66 vs. 54.46), it achieved comparable, or even better, reconstruction quality when paired with the same reconstruction method ( Fiber-F1: 45.7 vs. 46.8 vs. 45.6).
Moreover, when clDice was used as the segmentation model, neuTube, APP2, and Kimimaro all achieved consistently strong performance, whereas the strong performance of SegMamba and IVNet depended more heavily on the choice of reconstruction algorithm.
This suggests that reconstruction algorithms are affected by distinct error patterns produced by different segmentation models, indicating a complex coupling between segmentation method and reconstruction method.

\paragraph{Geometry-based metrics are not suitable for evaluating the quality of neuron reconstruction.}
Methods that perform well in terms of local geometric reconstruction do not necessarily achieve better structural reconstruction quality.
For example, Kimimaro obtained the best SSD, Point-F1, Length-F1, and Leaf-F1 scores across multiple segmentation models, but its Bifurcation-F1 and Fiber-F1 were substantially lower than those of other methods.
In contrast, neuTube and APP2 achieved only moderate performance on geometry-based metrics, yet performed strongly on Bifurcation-F1 and Fiber-F1.
Figure~\ref{fig:res_compare_metrics} shows the correlations among different metrics computed from our large-scale experimental results. From a statistical perspective, geometry-based metrics exhibit certain correlations with each other, but show no clear correlation with structure-based metrics.
\paragraph{Learning-based end-to-end reconstruction algorithms have great potential.}

It provides a potential path toward addressing the optimization difficulties caused by the coupling between segmentation and reconstruction methods in two-stage pipelines.
However, existing learning-based methods, such as SPE-DNR and NETracer, mainly predict local neurite information and show limited ability to reconstruct long-range neurites, as indicated by their low Fiber-F1 scores.
In contrast, the strong performance of SegMamba, which explicitly incorporates long-range context modeling, further demonstrates the importance of modeling long-range information for tracing long neurites.

\subsection{Brain-wide Reconstruction}
\begin{figure*}[t]
    \centering
    \includegraphics[width=\textwidth]{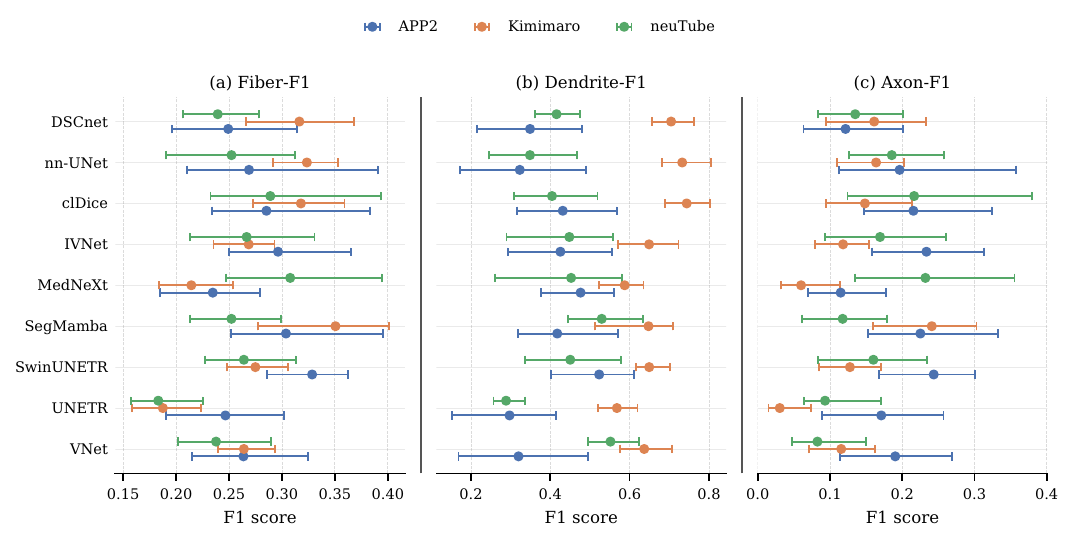}
    \caption{Whole-brain reconstruction performance across nine segmentation methods and three reconstruction methods. Points denote micro-F1 estimates and horizontal whiskers denote 95\% BCa bootstrap confidence intervals based on 10,000 neuron-resampling replicates.}
    \label{fig:wb_results}
\end{figure*}

We selected several representative methods by jointly considering their reconstruction performance and efficiency, and evaluated them at the brain-wide scale through our proposed whole-brain tracing framework.
Because tracing speed varies substantially across methods and neuron sizes differ greatly, the runtime for reconstructing a single neuron ranged from 30 seconds to 13.35 hours. In total, evaluating all methods required more than 400 GPU-hours.
The results are shown in Figure~\ref{fig:wb_results}.

As shown in the figure, clDice with Kimimaro achieves the best performance in dendrite reconstruction, whereas SwinUNETR with APP2 performs best in axon reconstruction.
Although SegMamba with Kimimaro achieves the best overall performance, its Fiber-F1 remains only 35.1\%, highlighting the substantial gap that still exists toward fully automated reconstruction. While existing methods can reconstruct dendrites reasonably well, they still struggle to reconstruct long-range axonal structures, highlighting the challenge of long-distance continuity in brain-wide reconstruction.
Kimimaro achieves better reconstruction quality for dendrites, whereas no reconstruction method demonstrates a significant advantage in axon reconstruction.
The relative performance of methods also differs between dendrite and axon reconstruction, suggesting that strong performance on one structure does not necessarily translate into strong performance on the other.

\begin{figure}[htbp]
    \centering
    \includegraphics[width=0.98\linewidth]{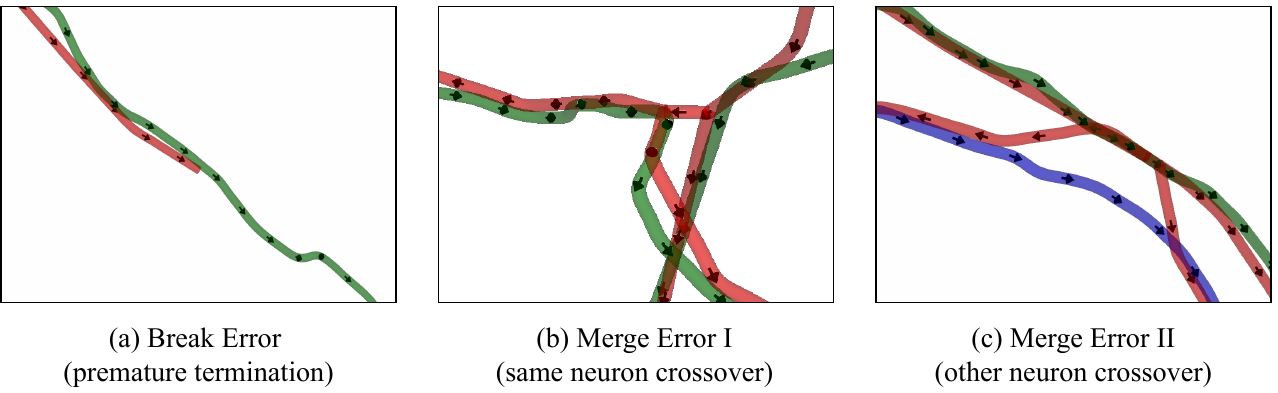}
    \caption{Visualization of reconstruction errors in brain-wide neuron reconstruction. The ground-truth annotations of neuron 1 and neuron 2 are shown in green and blue, respectively, while the prediction for neuron 1 is shown in red. The arrows along the tubular structures indicate the fiber direction.}
    \label{fig:bad_case}
\end{figure}

Figure~\ref{fig:bad_case} shows the representative errors that occur in brain-wide reconstruction. As illustrated in the figure, errors in brain-wide neuron reconstruction mainly fall into three categories.
\textbf{(1) Break error:}
the predicted fiber terminates prematurely and fails to follow the correct continuation in the ground truth, leading to missing branches or incomplete long-range projections.
\textbf{(2) Merge error I (same-neuron crossover):}
two nearby fibers from the same neuron are incorrectly connected, which alters the intrinsic branching organization of the target neuron.
\textbf{(3) Merge error II (other-neuron crossover):}
the tracing result incorrectly crosses from the target neuron to a different neuron, causing the predicted reconstruction of neuron 1 to erroneously absorb fibers belonging to neuron 2.

These three error types can have a substantial impact on automatic brain-wide neuron reconstruction: a single local error may propagate and affect the reconstruction at the whole-brain scale. For example, (1) a Break Error near the proximal axon can cause large range axonal branches to be missed; (2) Merge Error I may cause the tracing process to jump onto another fiber and trace backward toward the soma, producing many false positives and false negatives; and (3) Merge Error II may lead the reconstruction to follow another neuron, generating numerous false positives while also wasting tracing time.
These errors highlight the central challenge of brain-wide neuron reconstruction: accurate local geometry alone is insufficient for structurally correct reconstruction. A reliable method must preserve continuity, branching structure, and neuronal identity over long distances.

\section{Conclusion}
We introduced CORAL, the first benchmark for automatic brain-wide neuron reconstruction from light microscopy images.
By unifying block-level and brain-wide reconstruction tasks, the structure-aware metric, and the whole-brain neuron tracing Framework, CORAL provides a standardized platform for developing and testing brain-wide neuron reconstruction methods.
We conducted extensive experiments on this benchmark and identified several key insights that may help guide future research in automated neuron reconstruction.
We hope CORAL will serve as an efficient platform for future research on brain-wide neuron reconstruction.

\begin{ack}
This work was supported in part by the Fundamental and Interdisciplinary Disciplines Breakthrough Plan of the Ministry of Education of China (No. JYB2025XDXM504), the National Natural Science Foundation of China (No. 62576187) and the Guoqiang Institute, Tsinghua University.
\end{ack}

\bibliographystyle{unsrt}
\bibliography{main}

\clearpage
\appendix
\setcounter{table}{0}
\renewcommand{\thetable}{A.\arabic{table}}
\begin{center}
{\Large\bfseries Contents of Appendix}
\end{center}
\startcontents[appendix]
\printcontents[appendix]{}{1}{\setcounter{tocdepth}{2}}
\newpage
\section{SWC Format and Evaluation Metrics}
\label{sec:metric}
This section defines the representation and reconstruction metrics used in our benchmark. The definitions follow our evaluation implementation. We denote the ground-truth and predicted reconstructions by $T_g$ and $T_p$, respectively. When required, a scale vector $\mathbf{s}=(s_x,s_y,s_z)$ is applied coordinate-wise to transform both reconstructions into a common coordinate system.

\subsection{SWC Format}
We use the SWC format~\cite{swc-format} to represent neuronal morphology as a rooted tree, or a forest when a local block contains multiple disconnected components, embedded in $\mathbb{R}^3$. Each row of an SWC file describes one sampled skeleton node as a seven-tuple
\[
v_i=(i,t_i,x_i,y_i,z_i,r_i,p_i),
\]
where $i$ is the node identifier, $t_i$ is its anatomical type, $(x_i,y_i,z_i)$ are its spatial coordinates, $r_i$ is the local radius, and $p_i$ is the identifier of its parent. The conventional type codes $1$, $2$, $3$, and $4$ denote soma, axon, basal dendrite, and apical dendrite, respectively. A node with $p_i=-1$ is a root. Every non-root node and its parent form a straight edge in three-dimensional space; hence, the parent field induces a directed tree structure. In a complete reconstruction the root generally corresponds to the soma, whereas a component clipped by a block boundary may have an artificial root.

For the geometric metrics below, let $R_h(T)$ denote the topology-preserving resampling used in our implementation: each chain between consecutive key nodes (roots, bifurcations, or leaves) is resampled at approximately uniform spacing $h$, while the key nodes are retained. For two finite point sets $X$ and $Y$, define the nearest-neighbor distance
\[
d(\mathbf{x},Y)=\min_{\mathbf{y}\in Y}\|\mathbf{x}-\mathbf{y}\|_2.
\]

\subsection{Spatial Distance (SD)}
Spatial distance measures the average geometric discrepancy between two reconstructions~\cite{peng2010automatic}. In the implementation, both trees are first scaled and then resampled using $h_s$, which is also used as the substantial-distance threshold. Let $X_g$ and $X_p$ be the node-coordinate sets of $R_{h_s}(\mathbf{s}\odot T_g)$ and $R_{h_s}(\mathbf{s}\odot T_p)$. The two directed distances are
\[
\mathrm{SD}_{g\rightarrow p}=\frac{1}{|X_g|}
\sum_{\mathbf{x}\in X_g}d(\mathbf{x},X_p),\qquad
\mathrm{SD}_{p\rightarrow g}=\frac{1}{|X_p|}
\sum_{\mathbf{x}\in X_p}d(\mathbf{x},X_g),
\]
and the reported score is
\[
\mathrm{SD}=\frac{\mathrm{SD}_{g\rightarrow p}+\mathrm{SD}_{p\rightarrow g}}{2}.
\]
The directed terms reflect geometric omission and commission errors, respectively. A smaller SD indicates closer overall spatial agreement. We use $h_s=2$ by default.

\subsection{Substantial Spatial Distance (SSD)}
SSD suppresses small localization deviations and averages only nearest-neighbor distances strictly larger than $h_s$~\cite{peng2010automatic,PyNeval}. Using the same resampled sets as SD, define
\[
\mathcal{S}_{g\rightarrow p}=\{\mathbf{x}\in X_g:d(\mathbf{x},X_p)>h_s\},
\qquad
\mathcal{S}_{p\rightarrow g}=\{\mathbf{x}\in X_p:d(\mathbf{x},X_g)>h_s\}.
\]
For $(a,b)\in\{(g,p),(p,g)\}$, the directed SSD is
\[
\mathrm{SSD}_{a\rightarrow b}=
\begin{cases}
\displaystyle\frac{1}{|\mathcal{S}_{a\rightarrow b}|}
\sum_{\mathbf{x}\in\mathcal{S}_{a\rightarrow b}}d(\mathbf{x},X_b),
& |\mathcal{S}_{a\rightarrow b}|>0,\\[6pt]
0, & \text{otherwise},
\end{cases}
\]
and
\[
\mathrm{SSD}=\frac{\mathrm{SSD}_{g\rightarrow p}
+\mathrm{SSD}_{p\rightarrow g}}{2}.
\]
Thus, SSD describes the magnitude of substantial errors rather than their frequency; lower is better. The implementation also computes the proportion of substantially different nodes,
\[
\mathrm{SSD}_{\%}=
\frac{|\mathcal{S}_{g\rightarrow p}|+|\mathcal{S}_{p\rightarrow g}|}
{|X_g|+|X_p|}.
\]

\subsection{Point Metrics}
Point metrics measure bidirectional nearest-node coverage~\cite{ADTL-Net}. In our implementation, each tree is resampled with step $h_p$ and the resulting coordinates are then scaled. Let $Y_g$ and $Y_p$ denote these two node sets. A predicted point and a ground-truth point are counted as covered when their respective nearest-neighbor distances are strictly smaller than $\tau_p$:
\[
\mathcal{C}_p=\{\mathbf{x}\in Y_p:d(\mathbf{x},Y_g)<\tau_p\},\qquad
\mathcal{C}_g=\{\mathbf{y}\in Y_g:d(\mathbf{y},Y_p)<\tau_p\}.
\]
The point precision, recall, and F1 score are therefore
\[
P_{\mathrm{point}}=\frac{|\mathcal{C}_p|}{|Y_p|},\qquad
R_{\mathrm{point}}=\frac{|\mathcal{C}_g|}{|Y_g|},\qquad
\mathrm{F1}_{\mathrm{point}}=\frac{2P_{\mathrm{point}}R_{\mathrm{point}}}
{P_{\mathrm{point}}+R_{\mathrm{point}}}.
\]
Let $N_{\mathrm{miss}}=|Y_g|-|\mathcal{C}_g|$ and $N_{\mathrm{extra}}=|Y_p|-|\mathcal{C}_p|$ be the numbers of missed and extra points. The miss-extra score (MES) is
\[
\mathrm{MES}=\frac{|Y_g|-N_{\mathrm{miss}}}
{|Y_g|+N_{\mathrm{extra}}}.
\]
The nearest-neighbor associations are directional and need not be one-to-one. If either point set is empty, our implementation returns zero for all four point scores. We use $h_p=2$ and $\tau_p=4$ by default.

\subsection{Length-F1}
Length-F1 evaluates matched edge length and incorporates local connectivity in addition to point proximity~\cite{PyNeval}. Both trees are resampled with step $h_l$ and then scaled. For every ground-truth edge $e_g=(u,v)$, we find candidate predicted edges near $u$ and $v$ and project the endpoints to the candidates. Let $\hat u$ and $\hat v$ be the projected points and $P_p(\hat u,\hat v)$ the unique path between them in the predicted tree. The ground-truth edge is accepted when
\[
\max\{\operatorname{dist}(u,e_p^u),\operatorname{dist}(v,e_p^v)\}\leq\tau_l,\qquad
\big|L(P_p(\hat u,\hat v))-L(e_g)\big|<\rho_l L(e_g),
\]
where $\operatorname{dist}(\cdot,e)$ is Euclidean point-to-segment distance, and no part of the predicted path has already been assigned to another ground-truth edge. This last constraint prevents duplicated credit. Let $M_g$ be the set of accepted ground-truth edges and let $L_p^{\mathrm{match}}$ be the sum of the corresponding predicted path lengths. Then
\[
P_{\mathrm{length}}=\frac{L_p^{\mathrm{match}}}{L(T_p)},\qquad
R_{\mathrm{length}}=\frac{L(M_g)}{L(T_g)},\qquad
\mathrm{F1}_{\mathrm{length}}=\frac{2P_{\mathrm{length}}R_{\mathrm{length}}}
{P_{\mathrm{length}}+R_{\mathrm{length}}+\epsilon}.
\]
We use $h_l=2$, $\tau_l=2$, $\rho_l=0.2$, and the numerical guard $\epsilon=10^{-6}$. Compared with Point-F1, Length-F1 rejects spatially close structures whose connectivity or intervening path length is inconsistent.

\subsection{Keypoint Metrics}
We define two classes of topologically important nodes. A \emph{bifurcation} is a non-root node with more than one child, and a \emph{leaf} is a node with no children. In block-level evaluation, an artificial root with exactly one child is additionally treated as a leaf because it represents a neurite truncated by the block boundary; this rule is disabled for whole-brain evaluation.

Let $\mathcal{K}_g$ and $\mathcal{K}_p$ be the scaled coordinate sets of the selected ground-truth and predicted keypoint types. We compute their pairwise Euclidean distances and use the Hungarian algorithm to obtain the minimum-cost one-to-one assignment $\mathcal{M}_k$. An assigned pair is a true positive only when its distance is strictly smaller than $\tau_k$:
\[
\mathrm{TP}_k=\sum_{(\mathbf{x},\mathbf{y})\in\mathcal{M}_k}
\mathbb{I}\!\left(\|\mathbf{x}-\mathbf{y}\|_2<\tau_k\right).
\]
With $\mathrm{FP}_k=|\mathcal{K}_p|-\mathrm{TP}_k$ and $\mathrm{FN}_k=|\mathcal{K}_g|-\mathrm{TP}_k$, the scores are
\[
P_{\mathrm{key}}=\frac{\mathrm{TP}_k}{|\mathcal{K}_p|},\qquad
R_{\mathrm{key}}=\frac{\mathrm{TP}_k}{|\mathcal{K}_g|},\qquad
\mathrm{F1}_{\mathrm{key}}=\frac{2P_{\mathrm{key}}R_{\mathrm{key}}}
{P_{\mathrm{key}}+R_{\mathrm{key}}}.
\]
The main keypoint score uses the union of bifurcations and leaves. We additionally repeat the same matching independently for each class and report Bifurcation-Precision/Recall/F1 and Leaf-Precision/Recall/F1. We use $\tau_k=5$.

\subsection{DIADEM}
For reproducibility, we define the DIADEM value exactly as computed by our evaluation implementation. Each non-root edge of a scaled tree is divided into $n_e=\lceil L(e)/h_d\rceil$ equal subsegments. Their midpoints form a sample set $Q(T)$, and every midpoint $\mathbf{q}$ has weight $w(\mathbf{q})=L(e)/n_e$. A ground-truth sample is covered if its nearest predicted sample is within $\tau_d$, and vice versa:
\[
\mathcal{H}_g=\{\mathbf{q}\in Q(T_g):d(\mathbf{q},Q(T_p))\leq\tau_d\},
\qquad
\mathcal{H}_p=\{\mathbf{q}\in Q(T_p):d(\mathbf{q},Q(T_g))\leq\tau_d\}.
\]
The length-weighted recall, precision, and reported score are
\[
R_d=\frac{\sum_{\mathbf{q}\in\mathcal{H}_g}w(\mathbf{q})}
{\sum_{\mathbf{q}\in Q(T_g)}w(\mathbf{q})},\qquad
P_d=\frac{\sum_{\mathbf{q}\in\mathcal{H}_p}w(\mathbf{q})}
{\sum_{\mathbf{q}\in Q(T_p)}w(\mathbf{q})},\qquad
\mathrm{DIADEM}=\frac{2P_dR_d}{P_d+R_d}.
\]
We use $h_d=1$ and $\tau_d=2$. If both trees have zero total edge length, the score is one; if only one is empty, it is zero. This implementation is a length-weighted bidirectional coverage score and should be distinguished from the classical ancestor-aware DIADEM protocol~\cite{DIADEMMetric,PyNeval}.

\subsection{Fiber Metrics}
Given a rooted SWC tree $T$, a \emph{fiber} is the unique root-to-leaf path
\[
f=(v_{i_1},v_{i_2},\ldots,v_{i_K}),
\]
where $v_{i_1}$ is a root, $v_{i_K}$ is a leaf, and each node is the parent of the next. Its length is
\[
L(f)=\sum_{k=1}^{K-1}\|\mathbf{x}_{i_{k+1}}-\mathbf{x}_{i_k}\|_2.
\]
Each input tree is first resampled with step $h_0=2$. For block-level evaluation, each predicted component is re-rooted at the node nearest to a ground-truth root whenever their distance in the input coordinate system is at most $20$; this operation changes the root and edge directions but does not translate coordinates. The coordinates are then scaled and root-to-leaf paths are extracted.

For FIoU computation, each fiber is again resampled along arc length with step $h_f$. Let $Z_f=(\mathbf{z}_1,\ldots,\mathbf{z}_m)$ be this ordered point sequence, $\mathbf{c}_i=(\mathbf{z}_i+\mathbf{z}_{i+1})/2$ its $i$th segment midpoint, and $\delta_i=\|\mathbf{z}_{i+1}-\mathbf{z}_i\|_2$ the corresponding segment length. The directed overlap is
\[
O(f_a\!\rightarrow\!f_b)=\sum_{i=1}^{|Z_{f_a}|-1}
\mathbb{I}\!\left(d(\mathbf{c}_i,Z_{f_b})\leq\tau_f\right)\delta_i.
\]
The distance is thus evaluated against the resampled points of the other fiber. We symmetrize the overlap and define fiber intersection-over-union (FIoU) as
\[
O(f_a,f_b)=\frac{O(f_a\!\rightarrow\!f_b)+O(f_b\!\rightarrow\!f_a)}{2},\qquad
\mathrm{FIoU}(f_a,f_b)=\frac{O(f_a,f_b)}
{L(f_a)+L(f_b)-O(f_a,f_b)+\epsilon_{\mathrm{iou}}}.
\]
We use $h_f=1$, $\tau_f=5$, and $\epsilon_{\mathrm{iou}}=10^{-7}$. Fiber pairs with a length ratio below $0.5$ or a centroid distance above $100$ are pruned before exact overlap computation. In the direction-aware whole-brain setting, a pair is also pruned when the distance between its roots or between its leaves exceeds half of the longer fiber length.

Let $\mathcal{F}_p$ and $\mathcal{F}_g$ be the predicted and ground-truth fiber sets. We use the Hungarian algorithm to find the one-to-one assignment that maximizes total FIoU. If $\mathcal{M}_f$ denotes the assignment, a pair is correct only if its FIoU reaches $\eta$:
\[
\mathrm{TP}_{\mathrm{fiber}}=\sum_{(f_p,f_g)\in\mathcal{M}_f}
\mathbb{I}\!\left(\mathrm{FIoU}(f_p,f_g)\geq\eta\right).
\]
Accordingly,
\[
\mathrm{FP}_{\mathrm{fiber}}=|\mathcal{F}_p|-\mathrm{TP}_{\mathrm{fiber}},
\qquad
\mathrm{FN}_{\mathrm{fiber}}=|\mathcal{F}_g|-\mathrm{TP}_{\mathrm{fiber}},
\]
and the fiber precision, recall, and F1 score are
\[
P_{\mathrm{fiber}}=\frac{\mathrm{TP}_{\mathrm{fiber}}}{|\mathcal{F}_p|},\qquad
R_{\mathrm{fiber}}=\frac{\mathrm{TP}_{\mathrm{fiber}}}{|\mathcal{F}_g|},\qquad
\mathrm{F1}_{\mathrm{fiber}}=\frac{2P_{\mathrm{fiber}}R_{\mathrm{fiber}}}
{P_{\mathrm{fiber}}+R_{\mathrm{fiber}}+\epsilon_{\mathrm{f1}}}.
\]
We use $\epsilon_{\mathrm{f1}}=10^{-12}$ and set $\eta=0.8$ for block-level evaluation and $\eta=0.75$ for whole-brain evaluation. In the whole-brain setting, axon and dendrite scores are additionally obtained by repeating the assignment within fibers whose leaf types are $2$ and $3/4$, respectively. Unlike local geometric metrics, Fiber-F1 gives credit only when an entire root-to-leaf trajectory is recovered with sufficient spatial overlap.

\section{Whole-Brain Neuron Tracing Framework}
\label{sec:whole_brain_tracing_framework}

We design a local-to-global tracing framework for whole-brain neuron reconstruction from ultra-large 3D microscopy volumes.
The central idea is to decompose whole-brain reconstruction into an iterative local-to-global process. 
Specifically, the brain volume is partitioned into overlapping 3D cubes, and the reconstruction is grown from the soma by repeatedly performing three steps: \emph{search}, \emph{trace}, and \emph{grow}. The search module identifies candidate nodes for further expansion based on the current partial neuron tree; the tracing module predicts a local skeleton within the selected cube; and the growing module attaches the newly traced local structure to the candidate nodes. Repeating this procedure yields a whole-brain neuron tree assembled from cube-level predictions.

\subsection{Overlapping Cube Grid Construction}
To enable scalable tracing in an ultra-large brain volume, we organize the brain region of interest as an overlapping 3D grid of local cubes. Let the region of interest be denoted by $\Omega \subset \mathbb{R}^3$. We parameterize the grid by a cube size $(c_x, c_y, c_z)$ and a step size $(s_x, s_y, s_z)$, where typically $s_i < c_i$ for each axis so that adjacent cubes overlap with each other. This overlapping design allows the same neurite segment to be observed in multiple neighboring cubes, which is important for preserving local continuity across cube boundaries and for supporting reliable cross-cube merging.

Let $(x_0, y_0, z_0)$ denote the minimum coordinate of the brain region of interest. Each cube, indexed by $(i, j, k)$, corresponds to a cubic region defined as
\[
C_{i,j,k}
=
[x_0+i s_x,\; x_0+i s_x+c_x)
\times
[y_0+j s_y,\; y_0+j s_y+c_y)
\times
[z_0+k s_z,\; z_0+k s_z+c_z).
\]
Here, the cube origin is determined by the grid index and the step size, while the cube extent is determined by the fixed cube size. In this way, the entire brain volume is covered by a set of partially overlapping local subvolumes that serve as the basic processing units of our framework.

The grid plays two roles in our method. First, it provides a discrete search space for whole-brain exploration, allowing the reconstruction process to move from one local region to another in a structured manner rather than scanning the full volume exhaustively. Second, it defines the spatial support of each local tracing step, so that cube-level predictions can later be associated and merged into a global neuron tree. Because neighboring cubes share overlapping content, branches that pass through cube boundaries can still be matched based on their common local geometry, which substantially improves the robustness of local-to-global assembly.

\subsection{Local Cube Tracing}
\label{sec:local_tracing}
For each selected cube, we perform local neuron tracing within the corresponding image subvolume. Our framework is modular and can be readily coupled with different local tracing algorithms. This design makes the overall method flexible, as improvements in cube-level tracing can be directly incorporated into the whole-brain reconstruction pipeline.

Formally, for each cube $C_{i,j,k}$, we extract an image volume $I_{i,j,k}$ from the corresponding region of the whole-brain volume. The local tracing module then produces a cube-level skeleton
\[
S_{i,j,k} = \mathcal{T}(I_{i,j,k}),
\]
where $\mathcal{T}$ denotes the local tracing operator.
Since tracing is performed in the local coordinate system of the cube, the predicted skeleton is subsequently mapped back to the global coordinate system by adding the spatial offset of the cube origin.

To improve computational efficiency during iterative reconstruction, we cache the tracing result of each cube on disk. As a result, revisiting a previously processed cube does not require recomputing its local skeleton prediction. This significantly reduces the overall cost of the search--trace--merge procedure, particularly when multiple frontier nodes are associated with the same cube.

\subsection{Topology-aware Growing}
\label{sec:topology_growing}
To progressively construct a whole-brain neuronal morphology, we introduce a topology-aware growing module that incrementally integrates locally traced subtrees into a global reconstruction. Starting from a candidate frontier node, the module expands the global tree by identifying valid junctions and merging non-redundant structures in a topology-consistent manner.

\textbf{Grid-based localization.}
Given a candidate expansion node $p_c \in \mathbb{R}^3$ , we first determine the most relevant spatial region by selecting the cube whose center is closest to $p_c$:
$$
(i^*, j^*, k^*) = \arg\min_{i,j,k} \left\| \mathrm{center}(C_{i,j,k}) - p_c \right\|_2.
$$
This ensures that subsequent operations are performed within the most spatially relevant local context.

\textbf{Global-local subtree extraction.}
Within the selected cube $C_{i^*, j^*, k^*}$, we extract a local subtree $t_g \subset T$ from the current global reconstruction $T$, restricted to the cube region. The subtree is re-rooted at $p_c$, serving as the reference structure for subsequent matching.

\textbf{Candidate junction selection.}
To identify potential connection points, we select a set of candidate junction nodes from $S_{i^*, j^*, k^*}$. Specifically, we retrieve the $N_c$ nearest nodes to the node $p_c$ that belong to distinct connected components:
$$
\mathcal{P}_J = \{ p_j^{(1)}, \dots, p_j^{(N_c)} \}.
$$
This design avoids redundant candidates from the same subtree and improves robustness in complex branching regions.

\textbf{Fiber-level overlap matching.}
For each candidate junction node $p_j \in \mathcal{P}_J$, we consider its corresponding subtree $t_j$ (re-rooted at $p_j$) and decompose both $t_j$ and $t_g$ into sets of fibers (i.e., root-to-leaf paths).
We define an overlap-based matching criterion between a candidate fiber $f_1 \subset t_j$ and a reference fiber $f_2 \subset t_g$. Let $\mathrm{overlap}(f_1, f_2)$ denote their shared arc length. A valid junction is identified if:
$$
\frac{\mathrm{overlap}(f_1, f_2)}{\mathrm{length}(f_1)} > \tau
\quad \text{or} \quad
\frac{\mathrm{overlap}(f_1, f_2)}{\mathrm{length}(f_2)} > \tau,
$$
where $\tau$ is a predefined overlap threshold.
This criterion ensures that the candidate subtree shares sufficient structural consistency with the existing reconstruction.

\textbf{Conflict-aware merging.}
Once a valid junction is detected, we remove overlapping segments from $t_j$ to avoid duplication. Specifically, the overlapping portion of the fiber is detached by severing its parent connection, preserving only the novel structure.
The pruned subtree is then attached to the global tree $T$ as a child of $p_c$, provided that its total length exceeds a minimum threshold $L_{\min}$, which prevents the introduction of spurious short branches.

\subsection{Heuristic-guided Search}
\label{sec:heuristic_search}

To progressively explore the whole-brain volume and expand the reconstructed neuronal morphology, we propose a \emph{heuristic-guided search} strategy that iteratively grows a global tree from a set of frontier nodes. Starting from the soma annotation, an initial global tree $T$ is obtained via local tracing, and its frontier nodes are used to initialize a candidate set $\mathcal{Q}$.

At each iteration, a candidate node $p_c \in \mathcal{Q}$ is selected to guide further exploration. The search is driven by two types of structurally informative nodes: \emph{leafs}, which naturally indicate continuation directions of neuronal fibers, and \emph{margin nodes}, which lie close to the boundary of the currently explored region and encourage expansion into unvisited areas. A node $p \in T$ is considered a margin candidate if its distance to the boundary of the current cube $C$ satisfies
$$
d_{\mathrm{margin}}(p, C) \leq \tau',
$$
where $\tau'$ is a predefined threshold.

Given a selected node $p_c$, we perform local tracing (Section~\ref{sec:local_tracing}) and topology-aware merging (Section~\ref{sec:topology_growing}) to expand the global tree. Newly attached subtrees introduce additional frontier nodes, from which new candidates are extracted. Specifically, leaf nodes of the newly merged subtree are added to $\mathcal{Q}$, while internal nodes located near cube boundaries are included as margin candidates. To avoid redundant exploration, we maintain a visited set $\mathcal{V}$ and ensure that each node is processed at most once.

To improve robustness against error accumulation, we adopt a breadth-first search (BFS) strategy instead of depth-first search (DFS). This choice encourages more uniform spatial exploration and reduces the risk that early merging errors propagate along a single incorrect branch.

In addition, we employ a cube-level caching mechanism to improve efficiency and stability. For each spatial cube, previously traced results are reused; however, fibers that have already been incorporated into the global tree are explicitly removed from the cached results before reuse. This prevents duplicated structures and mitigates the reinforcement of erroneous connections.

Overall, the proposed search procedure can be summarized as an iterative expansion process:
$$
    \mathcal{Q} \leftarrow \text{Init}(T), \quad
    \textbf{while } \mathcal{Q} \neq \emptyset:
    \quad
    p_c \leftarrow \text{Pop}(\mathcal{Q}), \quad
    T \leftarrow \text{Expand}(T, p_c), \quad
    \mathcal{Q} \leftarrow \mathcal{Q} \cup \text{Update}(T).
$$
This heuristic-guided design balances exploration and reliability by integrating structural priors, traversal strategy, and redundancy control, enabling scalable and robust whole-brain reconstruction.

\section{Experiment settings}
\subsection{Segmentation Training and Testing Configuration}
\label{app:seg_training_config}
All segmentation models are trained under a unified protocol for fair comparison.
During training, we randomly crop 3D blocks of size $128 \times 128 \times 128$ from annotated cubes of size $300 \times 300 \times 300$.
For validation and testing, inference is performed on cubes of size $300 \times 300 \times 300$ using a sliding-window strategy with a window size of $128 \times 128 \times 128$ and an overlap ratio of 0.5.
We optimize the models using AdamW with a learning rate of $1\times10^{-4}$ and weight decay of $1\times10^{-4}$.
The loss function is Dice loss.
Training is conducted for 10,000 steps with a batch size of 4 on 4 GPUs using mixed-precision distributed training.
We use a linear warmup schedule at the beginning of training, followed by cosine annealing.
For preprocessing, training images are scaled to $[0,1]$, and random flipping is applied along the three spatial axes as data augmentation.
For validation and testing, we use percentile-based intensity normalization by mapping the 1st to 99th percentiles to $[0,1]$.
All experiments were conducted on four NVIDIA GeForce RTX 3090 GPUs.

\subsection{Implementation of Reconstruction Methods}
We make minor optimizations to the reconstruction methods implemented in Vaa3D (APP2, SmartTrace) to better suit our benchmark setting. The original implementations usually return only the brightest traced tree in a 3D volume. However, a local cube in our dataset may contain multiple disconnected neurite fragments. Therefore, we use an iterative trace-and-remove strategy: after each tracing step, the reconstructed tree is converted into a mask and removed from the image volume, and the tracer is then rerun on the residual image. The process stops when the returned SWC is empty or contains too few nodes.
All recovered SWC trees are finally merged into a single output.
This modification enables and to reconstruct multiple trees from one cube.

\section{Supplementary Experimental Results}

\subsection{Segmentation Results}
Table~\ref{tab:block_seg} presents the segmentation performance of ten different methods on the test set.
\begin{table}[htbp]
    \centering
    \caption{Comparison of different segmentation methods on the block-level task.}\label{tab:block_seg}
    \small
    \begin{tabular}{lccc|ccc}
    \toprule
    Method & Dice$\uparrow$ & clDice$\uparrow$ & NSD$\uparrow$ & \#Params(M)$\downarrow$ & GFLOPs$\downarrow$ & Latency(ms)$\downarrow$ \\
    \midrule
    VNet~\cite{VNet}              & 53.69 & 64.21 & 81.90 & 45.71  & 1011.39 & 52.92  \\
    nnU-Net~\cite{nnU-Net}         & 50.51 & 57.24 & 76.21 & 31.42  & 2141.14 & 33.29  \\
    UNETR~\cite{UNETR}            & 45.53 & 55.87 & 73.78 & 121.35 & 195.61  & 37.84  \\
    SwinUNETR~\cite{SwinUNETR}    & 50.40 & 61.12 & 78.60 & 15.51  & 199.68  & 92.01  \\
    MedNeXt~\cite{MedNeXt}        & 53.30 & 64.78 & 81.38 & 61.97  & 505.18  & 306.76 \\
    SegMamba~\cite{SegMamba}      & \textbf{54.66} & \underline{65.08} & \underline{82.13} & 64.24  & 1553.65 & 150.27 \\
    IVNet~\cite{IVNet} & \underline{54.46} & \textbf{65.14} & \textbf{82.45} & 53.62 & 1183.02 & 60.00 \\
    clDice~\cite{clDice}  & 48.04 & 63.21 & 72.25 & 31.42  & 2141.14 & 33.29  \\
    DSCNet~\cite{DSCNet}          & 53.07 & 64.01 & 81.41 & 4.23 & 423.73 & 178.18 \\
    ADTL-Net~\cite{ADTL-Net}       & 12.76 & 16.11 & 20.62 & 7.69   & 286.91  & 43.09  \\
    \bottomrule
    \end{tabular}
\end{table}

\subsection{Detailed Results on the Block-level Task}
\label{app:exp_detal_block}
Table~\ref{tab:detailed_block_level_results} presents the detailed quantitative results of different tracing methods on the block-level reconstruction task.
\begin{table*}[htbp]
\centering
\small
\setlength{\tabcolsep}{4pt}
\caption{Detailed block-level reconstruction performance of different methods. Best results are shown in bold, and second-best results are underlined.}\label{tab:detailed_block_level_results}
\resizebox{\linewidth}{!}{
\begin{tabular}{cccccccccc}
    \toprule
    \multicolumn{2}{c}{Method} & SD$\downarrow$ & SSD$\downarrow$ & Point-F1$\uparrow$ & MES$\uparrow$ & Length-F1$\uparrow$ & Leaf-F1$\uparrow$ & Bifurcation-F1$\uparrow$ & Fiber-F1$\uparrow$\\
    \midrule
    \multirow{9}{*}{APP2}& DSCNet & 13.6505 & 42.6313 & 0.8502 & 0.7636 & 0.8022 & 0.4082 & 0.4250 & 0.4082 \\
    & clDice & 14.5049 & 56.2745 & 0.8546 & 0.7546 & 0.8152 & 0.3319 & 0.3923 & 0.4581 \\
    & nn-UNet & 12.9215 & 41.1503 & 0.8608 & 0.7786 & 0.8136 & 0.4392 & 0.4353 & 0.4287 \\
    & IVNet & 13.3140 & 41.9819 & 0.8562 & 0.7727 & 0.8106 & 0.4190 & 0.4507 & 0.4483 \\
    & MedNeXt & 15.9024 & 48.6796 & 0.8439 & 0.7485 & 0.7922 & 0.3904 & 0.3975 & 0.3252 \\
    & SegMamba & 13.5240 & 43.9229 & 0.8574 & 0.7725 & 0.8122 & 0.4454 & 0.4662 & 0.4424 \\
    & SwinUNETR & 12.1924 & 41.6085 & 0.8520 & 0.7656 & 0.8031 & 0.3629 & 0.4408 & 0.3325 \\
    & UNETR & 11.4839 & 38.4829 & 0.8455 & 0.7472 & 0.7902 & 0.3107 & 0.4213 & 0.2875 \\
    & VNet & 12.6861 & 41.3826 & 0.8570 & 0.7766 & 0.8108 & 0.4257 & 0.4403 & 0.4395 \\
    \midrule
    \multirow{9}{*}{Kimimaro}& DSCNet & 10.8461 & 39.6433 & 0.8624 & 0.7763 & 0.8054 & 0.5771 & 0.3570 & 0.4210 \\
    & clDice & 9.8950 & 34.5246 & 0.8735 & 0.7924 & 0.8218 & 0.6103 & 0.3767 & 0.4487 \\
    & nn-UNet & 10.6941 & 39.7011 & 0.8717 & 0.7880 & 0.8189 & 0.5958 & 0.3818 & 0.4301 \\
    & IVNet & 10.7001 & 39.6400 & 0.8665 & 0.7841 & 0.8158 & 0.5706 & 0.3310 & 0.4109 \\
    & MedNeXt & 11.8287 & 44.5038 & 0.8649 & 0.7798 & 0.8050 & 0.5168 & 0.3338 & 0.3038 \\
    & SegMamba & 10.8848 & 39.9199 & 0.8618 & 0.7760 & 0.8114 & 0.5838 & 0.3979 & 0.4297 \\
    & SwinUNETR & 11.0525 & 31.7336 & 0.8433 & 0.7382 & 0.7743 & 0.4959 & 0.3378 & 0.3249 \\
    & UNETR & 14.8150 & 30.3950 & 0.7890 & 0.6472 & 0.7057 & 0.3923 & 0.2382 & 0.1904 \\
    & VNet & 10.9187 & 39.4200 & 0.8622 & 0.7766 & 0.8088 & 0.5659 & 0.3095 & 0.4063 \\
    \midrule
    \multirow{9}{*}{neuTube} & DSCNet & 16.4049 & 52.8636 & 0.8263 & 0.7214 & 0.7935 & 0.4689 & 0.4403 & 0.4369 \\
    & clDice & 13.4700 & 45.2117 & 0.8415 & 0.7572 & 0.8099 & 0.4939 & 0.4507 & 0.4574 \\
    & nn-UNet & 15.4649 & 51.4791 & 0.8345 & 0.7375 & 0.8027 & 0.4810 & 0.4446 & 0.4475 \\
    & IVNet & 16.8210 & 54.6989 & 0.8262 & 0.7221 & 0.7959 & 0.4771 & 0.4569 & 0.4563 \\
    & MedNeXt & 17.9398 & 56.7850 & 0.8188 & 0.7127 & 0.7871 & 0.4724 & 0.4250 & 0.4430 \\
    & SegMamba & 16.4829 & 54.9054 & 0.8313 & 0.7285 & 0.8028 & 0.4913 & 0.4906 & 0.4683 \\
    & SwinUNETR & 14.8432 & 48.1729 & 0.8237 & 0.7257 & 0.7899 & 0.4541 & 0.4208 & 0.4020 \\
    & UNETR & 15.7476 & 45.8373 & 0.7983 & 0.6805 & 0.7503 & 0.3659 & 0.2874 & 0.3064 \\
    & VNet & 16.4064 & 53.4323 & 0.8259 & 0.7250 & 0.7952 & 0.4732 & 0.4453 & 0.4449 \\
    \midrule
    \multirow{9}{*}{smartTrace} & DSCNet & 15.4830 & 42.2836 & 0.8213 & 0.7094 & 0.7876 & 0.3051 & 0.4303 & 0.3974 \\
    & clDice & 12.1445 & 34.3846 & 0.8106 & 0.7389 & 0.7795 & 0.3365 & 0.3870 & 0.4178 \\
    & nn-UNet & 14.8389 & 40.4214 & 0.8157 & 0.7152 & 0.7836 & 0.3041 & 0.4205 & 0.4021 \\
    & IVNet & 15.9178 & 42.6343 & 0.8189 & 0.7109 & 0.7873 & 0.3055 & 0.4388 & 0.4141 \\
    & MedNeXt & 18.7440 & 49.0162 & 0.7907 & 0.6646 & 0.7579 & 0.2781 & 0.4139 & 0.3910 \\
    & SegMamba & 15.8813 & 42.9193 & 0.8245 & 0.7193 & 0.7934 & 0.3251 & 0.4623 & 0.4092 \\
    & SwinUNETR & 14.1917 & 38.5577 & 0.8053 & 0.7141 & 0.7673 & 0.2903 & 0.4233 & 0.3951 \\
    & UNETR & 16.8839 & 36.5758 & 0.7834 & 0.6620 & 0.7391 & 0.2655 & 0.3973 & 0.3622 \\
    & VNet & 16.3725 & 43.0692 & 0.8237 & 0.7040 & 0.7923 & 0.3222 & 0.4424 & 0.4124 \\
    \midrule
    \multirow{9}{*}{SPE-DNR} & DSCNet & 16.3057 & 42.5090 & 0.8239 & 0.6949 & 0.7323 & 0.1966 & 0.0281 & 0.0713 \\
    & clDice & 34.7587 & 52.9042 & 0.5729 & 0.3415 & 0.5486 & 0.0657 & 0.0078 & 0.0309 \\
    & nn-UNet & 33.4240 & 52.2203 & 0.6036 & 0.3692 & 0.5694 & 0.0650 & 0.0096 & 0.0243 \\
    & IVNet & 15.1397 & 39.7491 & 0.8293 & 0.7026 & 0.7208 & 0.1769 & 0.0357 & 0.0565 \\
    & MedNeXt & 20.0328 & 48.3904 & 0.8074 & 0.6558 & 0.7073 & 0.1674 & 0.0332 & 0.0587 \\
    & SegMamba & 15.9594 & 42.3837 & 0.8293 & 0.6996 & 0.7316 & 0.1914 & 0.0332 & 0.0689 \\
    & SwinUNETR & 22.7519 & 45.9359 & 0.7519 & 0.5723 & 0.6445 & 0.1165 & 0.0251 & 0.0363 \\
    & UNETR & 27.2959 & 45.8865 & 0.6727 & 0.4646 & 0.5514 & 0.0756 & 0.0217 & 0.0182 \\
    & VNet & 14.7721 & 39.0651 & 0.8307 & 0.7051 & 0.7259 & 0.1812 & 0.0329 & 0.0593 \\
    \midrule
    \multicolumn{2}{c}{NETracer} & 28.2544 & 49.3877 & 0.6142 & 0.4291 & 0.4793 & 0.0245 & 0.0256 & 0.0320 \\
    \bottomrule
\end{tabular}}
\end{table*}

\subsection{Detailed Results on the Brain-wide Task}
Table~\ref{tab:whole_brain_f1} presents the detailed quantitative results of different tracing methods on the brain-wide reconstruction task, evaluated using fiber-based metrics.
\begin{table*}[t]
    \centering
    \caption{Whole-brain reconstruction results. Each value is the micro-F1 estimate followed by its 95\% BCa bootstrap confidence interval (10,000 resamples). The best and second-best estimates in each metric are shown in bold and underlined, respectively.}
    \label{tab:whole_brain_f1}
    \setlength{\tabcolsep}{3.5pt}
    \renewcommand{\arraystretch}{1.05}
    \scriptsize
    \begin{tabular}{clccc}
    \toprule
    \multirow{2}{*}{Reconstruction} & \multirow{2}{*}{Method} & \multicolumn{3}{c}{Micro-F1 (95\% BCa CI)} \\
    \cmidrule(lr){3-5}
    & & Fiber & Dendrite & Axon \\
    \midrule
    \multirow{9}{*}{APP2} & DSCnet & $0.249\;[0.196,\,0.314]$ & $0.349\;[0.216,\,0.479]$ & $0.122\;[0.063,\,0.201]$ \\
    & clDice & $0.285\;[0.234,\,0.384]$ & $0.431\;[0.317,\,0.567]$ & $0.216\;[0.147,\,0.324]$ \\
    & nn-UNet & $0.269\;[0.210,\,0.391]$ & $0.323\;[0.172,\,0.491]$ & $0.196\;[0.112,\,0.358]$ \\
    & IVNet & $0.296\;[0.250,\,0.365]$ & $0.425\;[0.293,\,0.555]$ & $0.234\;[0.159,\,0.313]$ \\
    & MedNeXt & $0.235\;[0.185,\,0.279]$ & $0.476\;[0.376,\,0.560]$ & $0.115\;[0.070,\,0.178]$ \\
    & SegMamba & $0.304\;[0.252,\,0.395]$ & $0.418\;[0.319,\,0.571]$ & $0.225\;[0.153,\,0.332]$ \\
    & SwinUNETR & $\underline{0.328}\;[0.286,\,0.362]$ & $0.523\;[0.402,\,0.610]$ & $\mathbf{0.244}\;[0.168,\,0.301]$ \\
    & UNETR & $0.247\;[0.190,\,0.302]$ & $0.297\;[0.152,\,0.414]$ & $0.171\;[0.089,\,0.257]$ \\
    & VNet & $0.264\;[0.215,\,0.325]$ & $0.320\;[0.168,\,0.495]$ & $0.190\;[0.114,\,0.269]$ \\
    \midrule
    \multirow{9}{*}{Kimimaro} & DSCnet & $0.316\;[0.266,\,0.368]$ & $0.705\;[0.657,\,0.763]$ & $0.161\;[0.095,\,0.233]$ \\
    & clDice & $0.318\;[0.273,\,0.359]$ & $\mathbf{0.744}\;[0.688,\,0.803]$ & $0.148\;[0.094,\,0.214]$ \\
    & nn-UNet & $0.324\;[0.291,\,0.353]$ & $\underline{0.733}\;[0.682,\,0.804]$ & $0.164\;[0.110,\,0.202]$ \\
    & IVNet & $0.269\;[0.235,\,0.293]$ & $0.649\;[0.571,\,0.723]$ & $0.118\;[0.080,\,0.154]$ \\
    & MedNeXt & $0.214\;[0.184,\,0.254]$ & $0.587\;[0.524,\,0.635]$ & $0.060\;[0.032,\,0.114]$ \\
    & SegMamba & $\mathbf{0.351}\;[0.277,\,0.402]$ & $0.648\;[0.512,\,0.708]$ & $\underline{0.241}\;[0.160,\,0.303]$ \\
    & SwinUNETR & $0.275\;[0.248,\,0.306]$ & $0.649\;[0.615,\,0.701]$ & $0.128\;[0.085,\,0.170]$ \\
    & UNETR & $0.188\;[0.159,\,0.223]$ & $0.568\;[0.521,\,0.620]$ & $0.031\;[0.015,\,0.074]$ \\
    & VNet & $0.264\;[0.240,\,0.293]$ & $0.637\;[0.576,\,0.707]$ & $0.116\;[0.071,\,0.162]$ \\
    \midrule
    \multirow{9}{*}{neuTube} & DSCnet & $0.239\;[0.206,\,0.278]$ & $0.416\;[0.361,\,0.476]$ & $0.135\;[0.084,\,0.201]$ \\
    & clDice & $0.289\;[0.233,\,0.393]$ & $0.404\;[0.309,\,0.519]$ & $0.217\;[0.124,\,0.379]$ \\
    & nn-UNet & $0.252\;[0.191,\,0.312]$ & $0.349\;[0.245,\,0.468]$ & $0.186\;[0.126,\,0.258]$ \\
    & IVNet & $0.267\;[0.213,\,0.331]$ & $0.448\;[0.290,\,0.558]$ & $0.169\;[0.093,\,0.261]$ \\
    & MedNeXt & $0.308\;[0.247,\,0.394]$ & $0.452\;[0.261,\,0.580]$ & $0.232\;[0.135,\,0.355]$ \\
    & SegMamba & $0.252\;[0.213,\,0.299]$ & $0.530\;[0.444,\,0.633]$ & $0.118\;[0.062,\,0.179]$ \\
    & SwinUNETR & $0.264\;[0.228,\,0.313]$ & $0.450\;[0.337,\,0.578]$ & $0.160\;[0.083,\,0.234]$ \\
    & UNETR & $0.183\;[0.157,\,0.225]$ & $0.288\;[0.257,\,0.337]$ & $0.093\;[0.064,\,0.171]$ \\
    & VNet & $0.238\;[0.202,\,0.290]$ & $0.552\;[0.496,\,0.624]$ & $0.083\;[0.047,\,0.150]$ \\
    \bottomrule
    \end{tabular}
\end{table*}

\subsection{Parameter Sensitivity of Fiber Metric}
We evaluated top 18 methods (ranked by performance at the threshold of 0.75) across IoU thresholds $\eta$. The results are presented in Table~\ref{tab:method_comparison}. The performance of all methods generally decreases as the IoU threshold increases, which is expected since stricter overlap requirements make it more challenging to achieve high scores. Notably, the relative ranking of methods remains largely consistent across different thresholds, indicating that the evaluation is robust to variations in $\eta$.

\begin{table*}[htbp]
    \centering
    \caption{Performance comparison under different IoU thresholds $\eta$. 
    The best results are shown in bold, and the second-best results are underlined.}
    \label{tab:method_comparison}
    \resizebox{\textwidth}{!}{%
    \begin{tabular}{lcccccccccc}
        \toprule
        Method
        & 0.50 & 0.55 & 0.60 & 0.65 & 0.70
        & 0.75 & 0.80 & 0.85 & 0.90 & 0.95 \\
        \midrule
        SegMamba+Kimimaro
        & \textbf{0.4024} & \textbf{0.3934} & \textbf{0.3818}
        & \textbf{0.3738} & \textbf{0.3638} & \textbf{0.3505}
        & \textbf{0.3405} & \textbf{0.3262} & \textbf{0.3124}
        & \textbf{0.2928} \\

        SwinUNETR+APP2
        & \underline{0.4016} & \underline{0.3905}
        & \underline{0.3781} & \underline{0.3631}
        & \underline{0.3432} & \underline{0.3285}
        & 0.3125 & 0.2975 & 0.2730 & 0.2567 \\

        nn-UNet+Kimimaro
        & 0.3626 & 0.3551 & 0.3487 & 0.3422 & 0.3326
        & 0.3235 & \underline{0.3171} & \underline{0.3096}
        & 0.3005 & 0.2904 \\

        clDice+Kimimaro
        & 0.3520 & 0.3445 & 0.3390 & 0.3331 & 0.3276
        & 0.3178 & 0.3130 & 0.3087
        & \underline{0.3038} & \underline{0.2924} \\

        DSCNet+Kimimaro
        & 0.3606 & 0.3515 & 0.3456 & 0.3348 & 0.3267
        & 0.3159 & 0.3094 & 0.3013 & 0.2927 & 0.2814 \\

        MedNeXt+neuTube
        & 0.3721 & 0.3610 & 0.3466 & 0.3356 & 0.3203
        & 0.3078 & 0.2952 & 0.2834 & 0.2690 & 0.2484 \\

        SegMamba+APP2
        & 0.3953 & 0.3810 & 0.3635 & 0.3433 & 0.3252
        & 0.3044 & 0.2876 & 0.2722 & 0.2554 & 0.2456 \\

        IVNet+APP2
        & 0.3667 & 0.3555 & 0.3432 & 0.3296 & 0.3110
        & 0.2960 & 0.2736 & 0.2485 & 0.2373 & 0.2262 \\

        clDice+neuTube
        & 0.3428 & 0.3323 & 0.3225 & 0.3124 & 0.3007
        & 0.2886 & 0.2781 & 0.2633 & 0.2500 & 0.2332 \\

        clDice+APP2
        & 0.3587 & 0.3458 & 0.3355 & 0.3190 & 0.3013
        & 0.2853 & 0.2700 & 0.2580 & 0.2432 & 0.2234 \\

        SwinUNETR+Kimimaro
        & 0.3150 & 0.3060 & 0.2998 & 0.2936 & 0.2845
        & 0.2749 & 0.2698 & 0.2619 & 0.2500 & 0.2359 \\

        IVNet+Kimimaro
        & 0.3116 & 0.2996 & 0.2921 & 0.2852 & 0.2766
        & 0.2686 & 0.2605 & 0.2554 & 0.2439 & 0.2341 \\

        nn-UNet+APP2
        & 0.3398 & 0.3284 & 0.3164 & 0.3001 & 0.2849
        & 0.2689 & 0.2474 & 0.2182 & 0.2057 & 0.1934 \\

        IVNet+neuTube
        & 0.3151 & 0.3068 & 0.2982 & 0.2903 & 0.2778
        & 0.2666 & 0.2570 & 0.2451 & 0.2297 & 0.2150 \\

        VNet+Kimimaro
        & 0.2976 & 0.2936 & 0.2826 & 0.2785 & 0.2716
        & 0.2640 & 0.2588 & 0.2542 & 0.2490 & 0.2403 \\

        SwinUNETR+neuTube
        & 0.3187 & 0.3088 & 0.2993 & 0.2891 & 0.2762
        & 0.2640 & 0.2492 & 0.2378 & 0.2249 & 0.2047 \\

        VNet+APP2
        & 0.3227 & 0.3140 & 0.3054 & 0.2902 & 0.2759
        & 0.2636 & 0.2482 & 0.2294 & 0.2163 & 0.2068 \\

        nn-UNet+neuTube
        & 0.3046 & 0.2953 & 0.2852 & 0.2762 & 0.2641
        & 0.2523 & 0.2437 & 0.2291 & 0.2136 & 0.1966 \\
        \bottomrule
    \end{tabular}%
    }
\end{table*}

\subsection{Generalization Performance on the Out-of-domain Test Set}
Table~\ref{tab:whole_brain_c166_f1} presents the performance of different models on the out-of-domain test set. C-166 is an independent whole-brain sample prepared and annotated following the same procedures and was used as the out-of-domain test set. It contains 15 neurons. All models were trained on the CORAL dataset and evaluated on C-166.
\begin{table*}[t]
    \centering
    \caption{Whole-brain reconstruction results on C-166 dataset. Each value is the micro-F1 estimate followed by its 95\% BCa bootstrap confidence interval (10,000 resamples). The best and second-best estimates in each metric are shown in bold and underlined, respectively.}
    \label{tab:whole_brain_c166_f1}
    \setlength{\tabcolsep}{3.5pt}
    \renewcommand{\arraystretch}{1.05}
    \scriptsize
    \begin{tabular}{clccc}
    \toprule
    \multirow{2}{*}{Reconstruction} & \multirow{2}{*}{Method} & \multicolumn{3}{c}{Micro-F1 (95\% BCa CI)} \\
    \cmidrule(lr){3-5}
    & & Fiber & Dendrite & Axon \\
    \midrule
    \multirow{8}{*}{APP2} & DSCnet & $0.155\;[0.082,\,0.243]$ & $0.331\;[0.195,\,0.484]$ & $0.084\;[0.027,\,0.162]$ \\
    & clDice & $\underline{0.296}\;[0.238,\,0.379]$ & $0.532\;[0.388,\,0.640]$ & $0.174\;[0.088,\,0.277]$ \\
    & nn-UNet & $0.285\;[0.230,\,0.379]$ & $0.535\;[0.361,\,0.649]$ & $0.162\;[0.087,\,0.272]$ \\
    & IVNet & $\mathbf{0.320}\;[0.259,\,0.420]$ & $0.505\;[0.344,\,0.637]$ & $\underline{0.216}\;[0.114,\,0.352]$ \\
    & MedNeXt & $0.022\;[0.008,\,0.045]$ & $0.072\;[0.030,\,0.128]$ & $0.001\;[0.000,\,0.009]$ \\
    & SegMamba & $0.263\;[0.217,\,0.332]$ & $0.506\;[0.395,\,0.591]$ & $0.168\;[0.086,\,0.260]$ \\
    & SwinUNETR & $0.211\;[0.150,\,0.291]$ & $0.232\;[0.181,\,0.296]$ & $0.192\;[0.096,\,0.323]$ \\
    & VNet & $0.253\;[0.203,\,0.346]$ & $0.500\;[0.318,\,0.642]$ & $0.104\;[0.036,\,0.267]$ \\
    \midrule
    \multirow{8}{*}{Kimimaro} & DSCnet & $0.050\;[0.019,\,0.118]$ & $0.160\;[0.071,\,0.319]$ & $0.005\;[0.001,\,0.015]$ \\
    & clDice & $0.224\;[0.170,\,0.293]$ & $\mathbf{0.578}\;[0.506,\,0.668]$ & $0.071\;[0.035,\,0.149]$ \\
    & nn-UNet & $0.143\;[0.103,\,0.206]$ & $0.392\;[0.295,\,0.488]$ & $0.038\;[0.019,\,0.091]$ \\
    & IVNet & $0.218\;[0.164,\,0.326]$ & $\underline{0.559}\;[0.483,\,0.647]$ & $0.077\;[0.034,\,0.169]$ \\
    & MedNeXt & $0.194\;[0.157,\,0.272]$ & $0.512\;[0.462,\,0.560]$ & $0.058\;[0.017,\,0.138]$ \\
    & SegMamba & $0.154\;[0.106,\,0.251]$ & $0.370\;[0.277,\,0.540]$ & $0.064\;[0.031,\,0.128]$ \\
    & SwinUNETR & $0.157\;[0.120,\,0.234]$ & $0.478\;[0.399,\,0.553]$ & $0.016\;[0.005,\,0.034]$ \\
    & VNet & $0.153\;[0.102,\,0.264]$ & $0.385\;[0.298,\,0.550]$ & $0.058\;[0.025,\,0.132]$ \\
    \midrule
    \multirow{8}{*}{neuTube} & DSCnet & $0.196\;[0.131,\,0.270]$ & $0.456\;[0.353,\,0.551]$ & $0.081\;[0.028,\,0.160]$ \\
    & clDice & $0.258\;[0.201,\,0.365]$ & $0.443\;[0.342,\,0.531]$ & $0.145\;[0.056,\,0.303]$ \\
    & nn-UNet & $0.230\;[0.160,\,0.317]$ & $0.374\;[0.208,\,0.550]$ & $0.113\;[0.041,\,0.211]$ \\
    & IVNet & $0.221\;[0.167,\,0.304]$ & $0.365\;[0.275,\,0.499]$ & $0.086\;[0.029,\,0.217]$ \\
    & MedNeXt & $0.291\;[0.201,\,0.437]$ & $0.366\;[0.241,\,0.521]$ & $\mathbf{0.240}\;[0.089,\,0.450]$ \\
    & SegMamba & $0.276\;[0.205,\,0.405]$ & $0.425\;[0.335,\,0.549]$ & $0.202\;[0.105,\,0.371]$ \\
    & SwinUNETR & $0.019\;[0.013,\,0.030]$ & $0.021\;[0.013,\,0.041]$ & $0.016\;[0.001,\,0.028]$ \\
    & VNet & $0.255\;[0.198,\,0.345]$ & $0.416\;[0.301,\,0.533]$ & $0.137\;[0.055,\,0.293]$ \\
    \bottomrule
    \end{tabular}
\end{table*}

\subsection{Performance of Individual Human Annotators}
To evaluate individual human performance in neuron annotation, we recruited five professional annotators. Each annotator was randomly assigned five neurons for annotation. Table~\ref{tab:human_annotation_metrics} reports their annotation performance for dendrites, axons, and the overall (Fiber) neuronal structure, along with the mean and standard deviation across the five annotators.
The results show relatively small inter-annotator variation, which is substantially lower than the performance gap between current state-of-the-art methods and human annotators. This indicates that existing algorithms remain far from achieving single-human-level performance.
\begin{table}[htbp]
    \centering
    \small
    \caption{Performance of Individual Human Annotators.}
    \label{tab:human_annotation_metrics}
    \resizebox{\textwidth}{!}{%
    \begin{tabular}{lccccccccc}
    \toprule
    Human & \multicolumn{3}{c}{Dendrite}
    & \multicolumn{3}{c}{Axon}
    & \multicolumn{3}{c}{Fiber} \\
    \cmidrule(lr){2-4}\cmidrule(lr){5-7}\cmidrule(lr){8-10}
    & Precision & Recall & F1
    & Precision & Recall & F1
    & Precision & Recall & F1 \\
    \midrule
    human1 & $0.897$ & $0.912$ & $0.901$
        & $0.746$ & $0.674$ & $0.708$
        & $0.958$ & $0.843$ & $0.895$ \\

    human2 & $0.993$ & $0.977$ & $0.984$
        & $0.935$ & $0.785$ & $0.849$
        & $0.946$ & $0.820$ & $0.875$ \\

    human3 & $0.845$ & $0.958$ & $0.857$
        & $0.758$ & $0.685$ & $0.720$
        & $0.967$ & $0.828$ & $0.887$ \\

    human4 & $0.976$ & $0.948$ & $0.961$
        & $0.966$ & $0.810$ & $0.880$
        & $0.970$ & $0.840$ & $0.900$ \\

    human5 & $0.853$ & $0.936$ & $0.867$
        & $0.758$ & $0.625$ & $0.682$
        & $0.965$ & $0.746$ & $0.830$ \\

    \midrule
    $\text{Mean}_{\text{Std}}$
    & $0.913_{0.069}$
    & $0.946_{0.025}$
    & $0.914_{0.057}$
    & $0.832_{0.108}$
    & $0.716_{0.078}$
    & $0.768_{0.090}$
    & $0.961_{0.010}$
    & $0.816_{0.040}$
    & $0.877_{0.028}$ \\
    \bottomrule
    \end{tabular}}
\end{table}

\section{Visualization of Bad Case in Brain-wide}
\begin{figure}[htbp]
    \centering
    \includegraphics[width=0.98\linewidth]{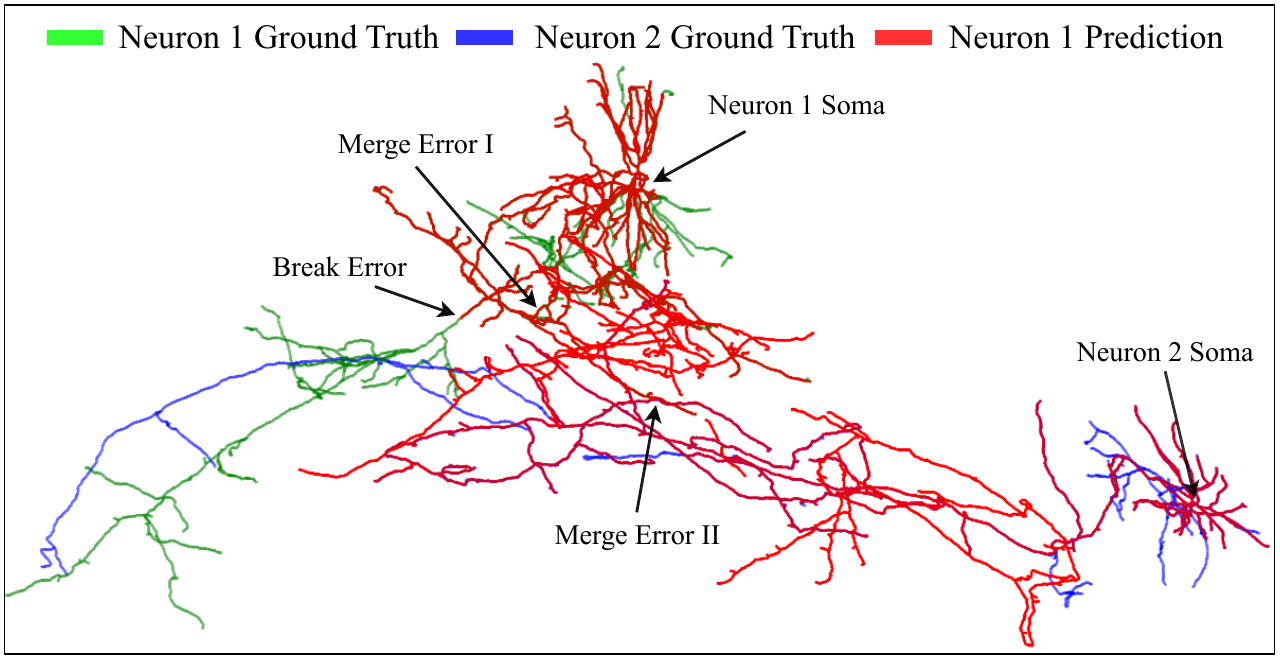}
    \caption{Visualization of reconstruction errors in brain-wide. The ground-truth annotations of neuron 1 and neuron 2 are shown in green and blue, respectively, while the prediction for neuron 1 is shown in red. The arrows along the tubular structures indicate the fiber direction.}
    \label{fig:bad_case_global}
\end{figure}

Figure~\ref{fig:bad_case_global} illustrates the impact of break and merge errors during whole-brain reconstruction. A break error in an axon causes multiple downstream neurites to be missed. Merge Error I leads the tracing process back toward the soma in the opposite direction, resulting in numerous false-positive predictions while leaving the affected region untraced. Merge Error II causes the reconstruction to erroneously trace into another neuron, producing a large number of false-positive predictions.

\section{Limitations}
Despite these strengths, CORAL has several limitations. The current benchmark is built from a limited number of fully annotated whole-brain neurons and therefore does not yet cover the full diversity of neuronal morphologies and imaging conditions. Moreover, all data are derived from mice, and the benchmark has not yet been evaluated on other species; thus, its cross-species generalizability remains to be established.
We view CORAL as a first step toward standardized brain-wide evaluation and expect future extensions to broaden data diversity, protocol coverage, and evaluation criteria.

\stopcontents[appendix]

\clearpage
\newpage
\section*{NeurIPS Paper Checklist}

\begin{enumerate}

\item {\bf Claims}
    \item[] Question: Do the main claims made in the abstract and introduction accurately reflect the paper's contributions and scope?
    \item[] Answer: \answerYes{}.
    \item[] Justification: The main claims made in the abstract and introduction accurately reflect the paper’s contributions and scope.

\item {\bf Limitations}
    \item[] Question: Does the paper discuss the limitations of the work performed by the authors?
    \item[] Answer: \answerYes{}.
    \item[] Justification: We discuss the limitations of the work in the appendix.

\item {\bf Theory assumptions and proofs}
    \item[] Question: For each theoretical result, does the paper provide the full set of assumptions and a complete (and correct) proof?
    \item[] Answer: \answerNA{}.
    \item[] Justification: The paper does not include theoretical results.

    \item {\bf Experimental result reproducibility}
    \item[] Question: Does the paper fully disclose all the information needed to reproduce the main experimental results of the paper to the extent that it affects the main claims and/or conclusions of the paper (regardless of whether the code and data are provided or not)?
    \item[] Answer: \answerYes{}.
    \item[] Justification: The paper fully discloses all the information needed to reproduce the main experimental results of the paper.

\item {\bf Open access to data and code}
    \item[] Question: Does the paper provide open access to the data and code, with sufficient instructions to faithfully reproduce the main experimental results, as described in supplemental material?
    \item[] Answer: \answerYes{}.
    \item[] Justification: The paper provide open access to the data and code.

\item {\bf Experimental setting/details}
    \item[] Question: Does the paper specify all the training and test details (e.g., data splits, hyperparameters, how they were chosen, type of optimizer) necessary to understand the results?
    \item[] Answer: \answerYes{}.
    \item[] Justification: The paper specify all the training and test details.

\item {\bf Experiment statistical significance}
    \item[] Question: Does the paper report error bars suitably and correctly defined or other appropriate information about the statistical significance of the experiments?
    \item[] Answer:  \answerYes{}.
    \item[] Justification: Error bars are not reported.

\item {\bf Experiments compute resources}
    \item[] Question: For each experiment, does the paper provide sufficient information on the computer resources (type of compute workers, memory, time of execution) needed to reproduce the experiments?
    \item[] Answer: \answerYes{}.
    \item[] Justification: The paper provide sufficient information on the computer resources.
    
\item {\bf Code of ethics}
    \item[] Question: Does the research conducted in the paper conform, in every respect, with the NeurIPS Code of Ethics \url{https://neurips.cc/public/EthicsGuidelines}?
    \item[] Answer: \answerYes{}.
    \item[] Justification: The research conducted in the paper conform, in every respect, with the NeurIPS Code of Ethics.

\item {\bf Broader impacts}
    \item[] Question: Does the paper discuss both potential positive societal impacts and negative societal impacts of the work performed?
    \item[] Answer: \answerNA{}.
    \item[] Justification: The paper is mainly aimed at promoting the advancement of neuroscience and will not have any negative social impacts.
    
\item {\bf Safeguards}
    \item[] Question: Does the paper describe safeguards that have been put in place for responsible release of data or models that have a high risk for misuse (e.g., pre-trained language models, image generators, or scraped datasets)?
    \item[] Answer: \answerNA{}.
    \item[] Justification: The paper poses no such risks.

\item {\bf Licenses for existing assets}
    \item[] Question: Are the creators or original owners of assets (e.g., code, data, models), used in the paper, properly credited and are the license and terms of use explicitly mentioned and properly respected?
    \item[] Answer: \answerYes{}.
    \item[] Justification: All the publicly available methods that were reimplemented have been appropriately cited.

\item {\bf New assets}
    \item[] Question: Are new assets introduced in the paper well documented and is the documentation provided alongside the assets?
    \item[] Answer: \answerYes{}.
    \item[] Justification: New assets introduced in the paper are well documented, and the documentation is provided alongside the assets.

\item {\bf Crowdsourcing and research with human subjects}
    \item[] Question: For crowdsourcing experiments and research with human subjects, does the paper include the full text of instructions given to participants and screenshots, if applicable, as well as details about compensation (if any)? 
    \item[] Answer: \answerNA{}.
    \item[] Justification: The paper does not involve crowdsourcing nor research with human subjects.

\item {\bf Institutional review board (IRB) approvals or equivalent for research with human subjects}
    \item[] Question: Does the paper describe potential risks incurred by study participants, whether such risks were disclosed to the subjects, and whether Institutional Review Board (IRB) approvals (or an equivalent approval/review based on the requirements of your country or institution) were obtained?
    \item[] Answer: \answerNA{}.
    \item[] Justification: The paper does not involve crowdsourcing nor research with human subjects.

\item {\bf Declaration of LLM usage}
    \item[] Question: Does the paper describe the usage of LLMs if it is an important, original, or non-standard component of the core methods in this research? Note that if the LLM is used only for writing, editing, or formatting purposes and does \emph{not} impact the core methodology, scientific rigor, or originality of the research, declaration is not required.
    \item[] Answer: \answerNA{}.
    \item[] Justification: The core method development in this research does not involve LLMs as any important, original, or non-standard components.

\end{enumerate}

\end{document}